\documentclass[letterpaper, 10 pt, conference]{ieeeconf}
\IEEEoverridecommandlockouts                            
\usepackage[T1]{fontenc} 
\usepackage{enumerate}
\usepackage{amsfonts}
\usepackage[colorinlistoftodos]{todonotes}
\usepackage[font=small,skip=12pt]{caption}
\usepackage{float}
\usepackage{subcaption}

\usepackage{booktabs}
\makeatletter
\let\NAT@parse\undefined
\makeatother
\usepackage{hyperref}
\usepackage{verbatim}
\usepackage{comment}
\usepackage{amsmath}
\usepackage{makecell}
\usepackage{mathtools}
\usepackage{dsfont}

\definecolor{ForestGreen}{rgb}{0.133, 0.545, 0.133}
\usepackage[ruled,vlined,linesnumbered]{algorithm2e}

\SetCommentSty{mycommfont}

\usepackage{xcolor,colortbl}

\definecolor{Gray}{gray}{0.85}
\definecolor{LightCyan}{rgb}{0.9,0.9,0.0}

\newcolumntype{a}{>{\columncolor{Gray}}c}
\newcolumntype{b}{>{\columncolor{LightCyan}}c}
\newcommand{\norm}[1]{\left\lVert#1\right\rVert}

\definecolor{cmtred}{rgb}{0.0,0.0,0.0}
\definecolor{cmtgreen}{rgb}{0.0,0.0,0.0}
\definecolor{cmtorange}{RGB}{0.0,0.0,0.0}

\title{\LARGE \bf
Differentiable Robotic Manipulation of
Deformable Rope-like Objects Using Compliant Position-based Dynamics
}

\author{
 Fei Liu$^{\dagger, 1}$ \IEEEmembership{Member, IEEE}, Entong Su$^{\dagger, 1}$, Jingpei Lu$^1$, Mingen Li$^1$ and Michael C. Yip$^1$ \IEEEmembership{Senior Member, IEEE}
\thanks{$\dagger$ These authors contributed equally. \protect\\ $^1$Advanced Robotics and Controls Lab, University of California San Diego, La Jolla, CA 92093 USA. {\tt\small \{f4liu, ensu, jil360, mil025, yip\}@ucsd.edu}}
}

\begin{document}

\maketitle 
\thispagestyle{empty}
\pagestyle{empty}

\begin{abstract}
\textcolor{cmtred}{Robot manipulation of rope-like objects is an interesting problem that has some critical applications, such as autonomous robotic suturing. Solving for and controlling rope is difficult due to the complexity of rope physics and the challenge of building fast and accurate models of deformable materials. While more data-driven approaches have become more popular for finding controllers that learn to do a single task, there is still a strong motivation for a model-based method that could be used to solve a large variety of optimization problems. Towards this end, we introduced compliant, position-based dynamics (XPBD) to model rope-like objects. Using geometric constraints, the model can represent the coupling of shear/stretch and bend/twist effects. Of crucial importance is that our formulation is differentiable, which can solve parameter estimation problems and improve the matching of rope physics to real-life scenarios (i.e., the real-to-sim problem). For the generality of rope-like objects, two different solvers are proposed to handle the inextensible and extensible effects of varied material stiffness for the rope. 
We demonstrate our framework's robustness and accuracy on real-to-sim experimental setups using the Baxter robot and the da Vinci research kit (DVRK) \cite{d2021accelerating}. Our work leads to a new path for robotic manipulation of the deformable rope-like object taking advantage of the ready-to-use gradients.
}

\end{abstract}
 
\section{Introduction}





For problems in grasping soft objects, modeling and simulation of deformable objects have recently been considerable interest in many robotic applications \cite{Huang_2021defgraspsim, Yunhai_2021}, to manipulating soft tissue \cite{Fei_2021, Yunhai_2021_PBD}, cloth \cite{Hoque_vsf_2020, Yilin_2020}, and even fluids \cite{Jingbin_2021, Schenck_2018_spnets}.  Among soft structures, deformable linear objects (DLOs), including rope-like objects, strings, cables, beams, etc., are studied (cable routing \cite{McConachie_2020}, wire insertion \cite{Lagneau_2020}, flexible rope \cite{Fangxun_2019}, and knotting of surgical thread \cite{Bo_Lu_2020}). Recently, techniques involving visual servoing \cite{Jihong_2020_Thesis, Jihong_2021}, latent dynamics learning \cite{Wenbo_Zhang_2021}, and adaptive estimation \cite{Mingrui_2021}, have all be explored for controlling DLOs. Further reviews in this space can be found in review papers on the topic \cite{Hang_Yin_2021, Sanchez_2018, Jihong_Zhu_2021}. 

Classical methods for modeling rope reside from Cosserat rod theory \cite{Gazzola_2018} which involves an analytical, partial differential equation representing the rope as a continuous 3D-dimensional curve exhibiting both bending and torsion. 
However, the analytical dynamics are limited by the computational efficiency and stability of solving this PDE with two-point boundary conditions (given by the start and end of the rope). 

Alternative to exact analytical models, a model-free method may try to identify representations of rope using purely visual descriptors from a camera without involving classical mechanics. These model-free representations include curvature and Fourier-based shapes \cite{Navarro_2014, Navarro_2018_TRO}, as well as neural network features \cite{Priya_2021_RSS, Shengzeng_2021, Angelina_2019}. 
Instead of directly modeling physical mechanics, the state space  representation is embedded in these latent space features. 
However, these approaches need a large quantity and variety of data for training, which is not easy to get in field robotic applications (either in simulation or real life). Meanwhile, the collected datasets are limited to the distribution of objects manipulated. Thus models that learn from simulation or limited real-world data encounter a real-to-sim transfer challenge or an over-fitting problem, respectively. 

\begin{figure}[t]
\centering
\includegraphics[width=1.0\linewidth]{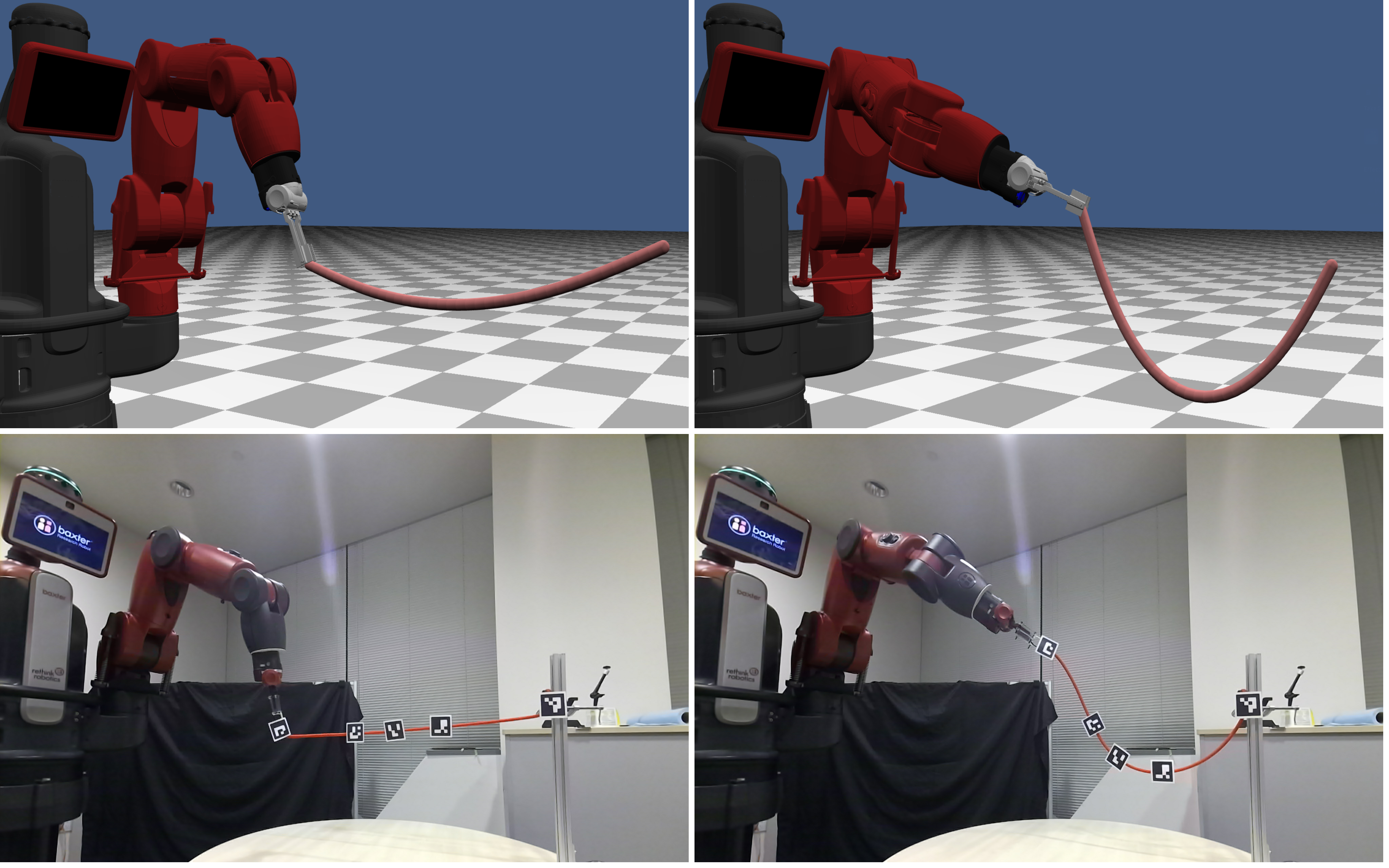}
\caption{\textcolor{cmtorange}{ \textbf{The physical experiments and simulation results for the shape control task of inextensible rope on Baxter} | The top row shows the simulation result for our shape control task of inextensible rope on the Baxter, rendered in the OpenGL. And the bottom rope is our physical result for this control task in the real scene. }}


\label{fig:Cover_Photo}
\end{figure}

An increasing number of simulators are made available for robot learning involving rope-like objects \cite{Rita_2021, Xingyu_corl2020softgym}. These simulators have been focused on their use for data generation and lack a method for integration in a model-based control context.
In general, 
many existing robotics simulators (including the above, as well as Bullet \cite{coumans_2021}) and Gym \cite{brockman2016openai}, unable to simulate soft bodies) only support gradient calculations based on finite differences. However, differentiable physics and simulation have become critical features for the robot learning community \cite{Difftaichi_2020}. The works in \cite{Avila_2018, Qiao_2021_Articulated, Werling_2021, Fei_2022parameter} have applied differentiable framework into rigid articulated body simulation, while others have done the same for deformable objects \cite{Krishna_2021_gradsim, Tao_2021_diffpd, Zhiao_Huang_2021}. These methods provide gradients within the differentiable framework, where optimization tasks for modeling and control are natively supported. Meanwhile, the differentiability can be easily deployed into neural layers for learning-based methods, such as in \cite{Qiao_2021_Multibody, PingchuanMa_2021diffaqua}.

One simulation approach that has gained significant interest due to its speed, stability, and capability to model the spectrum from rigid to soft to liquid is Position-based dynamics (PBD) \cite{Macklin_2017, Macklin_2014_Flex}. Unlike the traditional Lagrangian force-based method, the geometric position constraints are solved iteratively. Simulating soft objects involves utilizing the Compliant Position-based Dynamics (XPBD) method \cite{Macklin_2016_XPBD} which has been shown for a variety of deformable solid bodies, cloth, and chains. Some researchers have directly integrated the Cosserat model into PBD simulators to define the twist and bend, stretch and shear the a the constraints, by introducing particles with orientation information to describe angular updates \cite{Kugelstadt_2016_PBD}. Next, the authors added scale parameters for rope-like objects to preserve volume consistency \cite{Angles_2019_Viper}. \textcolor{cmtred}{Although several works have been applied for real-to-sim tasks of soft deformable tissue \cite{Fei_2021, Yunhai_2021_PBD}, rigid articulated robots \cite{Fei_2022}, and fluids \cite{Jingbin_2021, Florian_2021_CVPR}, with constraint-based formulation of PBD or XPBD.
However, there is a missing combination of modeling and control approaches applied to rope-like objects in literature, exploiting the capability of XPBD.
}

\textcolor{cmtred}{In the paper, we propose a solution to formulate a real-to-sim modeling and control framework.} We extended the original position-based dynamics (PBD) for rope-like objects proposed in \cite{Kugelstadt_2016_PBD} with a compliance parameter by following the XPBD work in \cite{Macklin_2016_XPBD}. Meanwhile, it is deployed differently using the automatic differentiation functionality available in PyTorch for gradient updates. The new compliant position-based dynamics (XPBD) for rope-like objects can guarantee stable forward simulation while the back-propagation of losses can be applied for modeling and control tasks and parameter identification. The contributions of this paper are as follows:
\begin{itemize}
    \item We describe rope-like objects as a  compliant position-based dynamics (XPBD) model by constructing geometrical constraints defining their behaviors.
    \item We introduced a differentiable framework for modeling and simulation that works well with auto-differentiation and algorithm configurations for constraint solving.
    \item \textcolor{cmtred}{We defined the problem of parameter identification and manipulation control of rope-like objects in real-to-sim context (Fig. \ref{fig:Cover_Photo}).
    \item We validate our methods in real-world experiments using robot manipulators (Baxter) and  surgical robots (dVRK).
    }
\end{itemize}
\section{Methods}
\label{sec:PBD_sim}




\subsection{Compliant Position-Based Dynamics (XPBD) Modeling of Rope-like Object}


\begin{figure}[!htb]
\begin{center}
\includegraphics[width=0.4\textwidth]{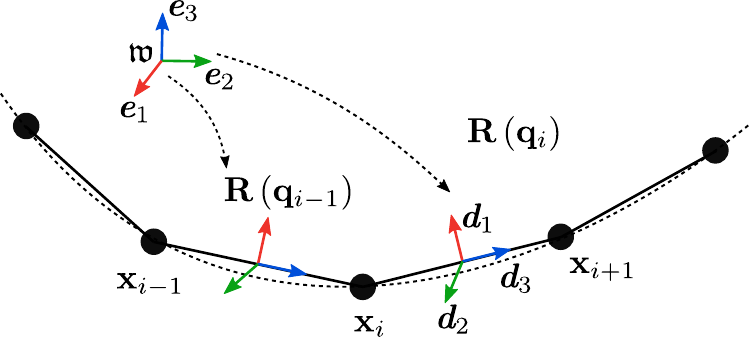}
\end{center}
\caption{\textbf{Proposed geometric model of rope-like objects} | The discrete particles include orientation representation of the rope-like object for modeling with compliant position-based dynamics (XPBD).}
\label{fig:pbd_rope_discretization}
\end{figure}

The XPBD method will be a foundation for building the differentiable model for DLOs. We firstly discretize the DLO into a sequence of particles (Fig. \ref{fig:pbd_rope_discretization}) with Cartesian coordinates $\mathbf{x} \in \mathbb{R}^3$. Meanwhile,  quaternions are used to describe orientations \textit{in-between} adjacent particles $\mathbf{q} = \left[ q_w, \mathbf{q}_v\right]\in \mathbf{SO}(3)$, and will be used to solve the bending and twist deformation of the DLO. Unlike force-based methods 
such as Euler-Bernoulli beam or Cosserat rod theory, full strain and torsion deformations can be updated with a position-based dynamics solver. As with all simulations using PBDs, the methods starts with a list of constraints, $\mathbf{C}(\mathbf{x} + \Delta\mathbf{x}, \mathbf{q} \oplus \Delta\mathbf{q})$ that describe the dynamics of particles. Solving the constraints involves updating $\Delta\mathbf{x}$ and $\Delta\mathbf{q}$ directly using a non-linear projected Jacobi method used for constrained optimization problems. The method to solve for the gradient updates is derived below.

Constraints can be linearized by Taylor series expansion,
\begin{equation}
{\small
\begin{split}
    & \mathbf{C}(\mathbf{x} + \Delta\mathbf{x}, \mathbf{q} \oplus \Delta\mathbf{q})  \\
    & \approx  
    \mathbf{C}(\mathbf{x},\mathbf{q}) + \nabla_{\mathbf{x}}\mathbf{C}(\mathbf{x},\mathbf{q}) \Delta\mathbf{x} + \nabla_{\mathbf{q}}\mathbf{C}(\mathbf{x},\mathbf{q}) \Delta\mathbf{q} \\
    & = 0 
\end{split}
}
\end{equation}
with
\begin{equation}
\label{eq:delta_positional_vector}
\begin{split}
    &  \Delta\mathbf{x}  = \mathbf{W}_\mathbf{x}\nabla^\top_\mathbf{x} \mathbf{C}(\mathbf{x}, \mathbf{q}) \Delta\boldsymbol{\Lambda} \\
    &  \Delta\mathbf{q}  = \mathbf{W}_{\mathbf{q}} \nabla^\top_{\mathbf{q}} \mathbf{C}(\mathbf{x}, \mathbf{q}) \Delta\boldsymbol{\Lambda} \\
\end{split}
\end{equation}
where the Lagrange multiplier change $\Delta\boldsymbol{\Lambda}$ can be found by introducing the compliance parameter $\boldsymbol{\alpha}$ \cite{Macklin_2016_XPBD}, 
\begin{equation}
\label{eq:Lagrange_multiplier_XPBD}
\begin{split}
    \displaystyle
    \Delta \boldsymbol{\Lambda} = - \left( \sum_{
    \substack{\pi \in  \{\mathbf{x}, \mathbf{q}\}}
    } \nabla_{\pi} \mathbf{C}  \mathbf{W}_{\pi} \nabla^T_{\pi} \mathbf{C} + \boldsymbol{\alpha} \right)^{-1} \left( \mathbf{C} + \boldsymbol{\alpha} \boldsymbol{\Lambda} \right)
\end{split}
\end{equation}

\noindent where $\mathbf{W}_{\mathbf{x}}$ and $\mathbf{W}_{\mathbf{q}}$ are the weighted terms to guarantee the conservation of linear and angular momentum. Generally, it refers to the mass/inertia matrix as $\mathbf{W}_{\mathbf{x}} = \mathrm{diag} \left(m^{-1}_1 \cdot \mathbf{1}, m^{-1}_2 \cdot \mathbf{1}, \cdots, m^{-1}_N \cdot \mathbf{1}\right)$ and $\mathbf{W}_{\mathbf{q}} = \mathrm{diag} \left( \mathbf{I}^{-1}_1, \mathbf{I}^{-1}_2, \cdots, \mathbf{I}^{-1}_N \right)$. For simplification of DLOs, we use uniformed scalar weights instead of matrices for each dimensionality, which has not much impact on simulation and dynamical performance but reduces computational load. That is, 
\begin{equation}
\begin{split}
& \mathbf{W}_{\mathbf{x}} = m^{-1}_{\mathbf{x}} ~~~ \mathbf{W}_{\mathbf{q}} = \mathbf{I}^{-1}_{\mathbf{q}} \\
\end{split}
\end{equation}


Next, we will introduce several geometrical constraints to simulate the behaviors of the DLOs.


\subsubsection{Shear and Stretch Constraint}
According to Cosserat theory, the shear and stretch measures the deformation regarding the tangent direction of the rope-like object. Therefore, the stretch/compressed length should be constrained relative its rest pose, which indicates in-extensible elasticity. Simultaneously, the normal direction (i.e., the rotated $e_3$ from world frame noted by $\mathfrak{w}$) for each cross-section should be parallel to the tangent direction of object's centerline, see Fig. \ref{fig:pbd_rope_discretization}. It measures the shear strain with respect to non-deformed states. Thus, for each pair of neighboring particles, the shear-stretch deformation can be integrated into a generalized constraint as,
$\mathbf{C}^{\mathcal{S}}(\mathbf{x}, \mathbf{q})=\left\{ \mathbf{c}_i^{\mathcal{S}}(\mathbf{x}_i, \mathbf{x}_{i+1}, \mathbf{q}_i)~|~i \in \left[1,2, \cdots, N-1\right] \right\}$, which is,
\begin{equation} 
\begin{split}
    \mathbf{c}_i^{\mathcal{S}}(\mathbf{x}_i, \mathbf{x}_{i+1}, \mathbf{q}_i) 
    & = \frac{\mathbf{x}_{i+1} - \mathbf{x}_{i}}{\left\lVert\mathbf{\bar{x}}_{i+1} - \mathbf{\bar{x}}_{i}\right\rVert} - \mathbf{R}\left(\mathbf{q}_{i}\right) \boldsymbol{\mathit{e}}_3
\end{split}
\label{eq:shear_stretch_const}
\end{equation}
where $\mathbf{R}\left(\mathbf{q}_{i}\right)$ is the rotation matrix from the local frame of $i^{th}$ line segment to world frame $\mathfrak{w}$ represented using quaternion, $\bar{\cdot}$ represent the states in rest pose. According to gradient calculation in \cite{Kugelstadt_2016_PBD}, we can obtain,
\begin{equation}
\begin{split}
    & \nabla^\top_{\mathbf{x}_{i}} \mathbf{c}_i^\mathcal{S} = - \frac{\mathbf{1}_{3 \times 3}}{\left\lVert\mathbf{\bar{x}}_{i+1} - \mathbf{\bar{x}}_{i}\right\rVert} \\
    & \nabla^\top_{\mathbf{x}_{i+1}} \mathbf{c}_i^\mathcal{S} = \frac{\mathbf{1}_{3 \times 3}}{\left\lVert\mathbf{\bar{x}}_{i+1} - \mathbf{\bar{x}}_{i}\right\rVert} \\
    & \nabla^\top_{\mathbf{q}_{i}} \mathbf{c}_i^\mathcal{S} = \nabla^\top_{\mathbf{q}_{i}} \left[ \mathbf{R}\left(\mathbf{q}_{i}\right) \boldsymbol{\mathit{e}}_3 \right] = \\
    & {\small 2 \left( q_{i, w} \boldsymbol{\mathit{e}}_3 - \boldsymbol{\mathit{e}}_3 \times \mathbf{q}_{i, v} | \mathbf{q}_{i, v}^\top \boldsymbol{\mathit{e}}_3 + \mathbf{q}_{i, v} \boldsymbol{\mathit{e}}^\top_3 - \boldsymbol{\mathit{e}}_3 \mathbf{q}^\top_{i, v} - \mathbf{q}^\top_{i, w} [\boldsymbol{\mathit{e}}_3]^{\times} \right)
    }
\end{split}
\end{equation}
Where $[\cdot]^\times$ is the skew-symmetrical matrix representation of a vector.

\subsubsection{Bend and Twist Constraint} In differential geometry, the Darboux vector $\mathbf{\Omega}$ is used to parameterize strain deformation with respect to frame rotation. According to Cosserat theory, the Darboux vector can be expressed as a quaternion by measuring the rod's twist in the tangent direction. Thus, the difference between the current and resting configuration should be evaluated, i.e. $\mathbf{\Omega} - \bar{\mathbf{\Omega}}$. According to \cite{Kugelstadt_2016_PBD}, the bend and twist constraint can be computed for each pair of two adjacent quaternions (shown in Fig. \ref{fig:pbd_rope_discretization}) by,
$\mathbf{C}^{\mathcal{B}}(\mathbf{q})=\left\{ \mathbf{c}_i^{\mathcal{B}}(\mathbf{q}_i, \mathbf{q}_{i+1})~|~i \in \left[1,2, \cdots, N-1\right] \right\}$, which is,
\begin{equation} 
\begin{split}
    \mathbf{c}_i^{\mathcal{B}}(\mathbf{q}_i, \mathbf{q}_{i+1}) & = \mathbf{\Omega} - \xi \cdot \bar{\mathbf{\Omega}} = \mathtt{Im}\left(\mathbf{q}^*_i \cdot \mathbf{q}_{i+1} - \bar{\mathbf{q}}^*_i \cdot \bar{\mathbf{q}}_{i+1} \right) \\
    \xi & = \mathtt{sign}(\mathbf{\Omega} + \bar{\mathbf{\Omega}})
\end{split}
\label{eq:bend_twist_const}
\end{equation}
where $\mathtt{Im}(\cdot)$ is the imaginary part of the quaternion and $(\cdot)^*$ is the conjugate quaternion. The constraint gradients can be calculated as, 
\begin{equation}
\begin{split}
    \nabla^\top_{\mathbf{q}_{i}} \mathbf{c}_i^\mathcal{B} & = -
    {\small \left( - \mathbf{q}_{i+1, v}~|~q_{i+1, w} \mathbf{1}_{3\times3}-[\mathbf{q}_{i+1, v}]^{\times} \right)
    } \\
    \nabla^\top_{\mathbf{q}_{i+1}} \mathbf{c}_i^\mathcal{B} & = +
    {\small \left( - \mathbf{q}_{i, v}~|~q_{i, w} \mathbf{1}_{3\times3}-[\mathbf{q}_{i, v}]^{\times} \right)
    }
\end{split}
\end{equation}

\subsubsection{Distance Constraint} One property of modeling ropes is that they can be considered either extensible or inextensible based on how stiff they are, and ultimately this can be modeled as a constraint. Although the above strain deformation have considered the inextensible property implicitly, the iterative solver in Eq. \ref{eq:delta_positional_vector} cannot guarantee that all the constraints will be satisfied. Therefore, we can explicitly consider the distance constraint for an enhanced inextensible chain structure using only discretized particle positions. We define $\mathbf{C}^{\mathcal{D}}(\mathbf{x})=\left\{ \mathbf{c}_i^{\mathcal{S}}(\mathbf{x}_i, \mathbf{x}_{i+1})~|~i \in \left[1,2, \cdots, N-1\right] \right\}$ by,
\begin{equation} 
\begin{split}
    \mathbf{c}_i^{\mathcal{D}}(\mathbf{x}_i, \mathbf{x}_{i+1}) 
    & = \left\lVert\mathbf{x}_{i+1} - \mathbf{x}_{i}\right\rVert - \left\lVert\mathbf{\bar{x}}_{i+1} - \mathbf{\bar{x}}_{i}\right\rVert
\end{split}
\label{eq:distance_const}
\end{equation}
The gradients can be easily obtained by,
\begin{equation}
\begin{split}
    - \nabla^\top_{\mathbf{x}_{i}} \mathbf{c}_i^\mathcal{D} = \nabla^\top_{\mathbf{x}_{i+1}} \mathbf{c}_i^\mathcal{D} = \frac{\mathbf{x}_{i+1} - \mathbf{x}_{i}}{\left\lVert\mathbf{\bar{x}}_{i+1} - \mathbf{\bar{x}}_{i}\right\rVert}
\end{split}
\end{equation}

\subsection{Real-to-Sim Parameter Identification} 
\label{sec:real2sim_parameters}

Given the above constraints, the dynamics of a rope-like object can be represented by a set of discretized particles (shown in Fig. \ref{fig:pbd_rope_discretization}) with position and orientation evolution. However, it is just an approximation of the real model for the rope objects. 

\textcolor{cmtred}{For the above shear/shear, bend/twist and distance constraints,} 
we can introduce the additional stiffness parameter to weight the updates during iteration steps, namely by, 
\begin{equation}
\begin{split}
   \boldsymbol{\eta}^* = \left\lbrace \boldsymbol{\eta}_{\mathbf{x}}^\mathcal{S}, \boldsymbol{\eta}_{\mathbf{q}}^\mathcal{S}, \boldsymbol{\eta}_{\mathbf{q}}^\mathcal{B}, \boldsymbol{\eta}_{\mathbf{x}}^\mathcal{D}, \boldsymbol{\eta}_{\mathbf{x}}^\mathcal{G}, \boldsymbol{\eta}^\mathcal{SOR}\right\rbrace
\end{split}
\end{equation}
Which stands for shear/stretch, bend/twist, and distance constraints regarding position or orientation. Moreover, $\boldsymbol{\eta}_{\mathbf{x}}^\mathcal{G}$ stands for position changes due to external gravity, and $\boldsymbol{\eta}^\mathcal{SOR}$ stands for the successive over-relaxation parameter and is applied to accelerate the convergence speed further. Thus, we can introduce the above parameters in a real-to-sim setup by inserting them into the following differentiable framework.

\subsection{Differential Framework} To perform an optimal control task, the ready-to-use gradients will be needed to compute. We introduce the compliant position-based dynamics (XPBD) inside a differentiable framework, as shown in Algorithm \ref{alg:differentiable_framework_XPBD_rope} \footnote{In this paper, we only consider the quasi-static dynamical states. Only the gravity will be regarded as without any external torques. Thus, the Euler prediction and integration of velocity and angular speed will not be involved.}. 
We rely on the automatic differentiation function provided by PyTorch to obtain the gradients. It can be natively integrated into optimization or learning methods, since the coding framework can be viewed as a differentiable layer which support both forward and backpropagation operations. The computing memory might be limited by the number of iterations for substep simulation. Denoting the gradient variable at time $t$ by $\boldsymbol{\lambda}^{t}$, we can formulate the following optimization problem as, 
\begin{equation}
\begin{split}
        \boldsymbol{\lambda}^{t}
    & = \operatorname*{argmin} \mathcal{L} \left( \mathbf{x}, \mathbf{q} \right) \\
        \operatorname*{s.t.} ~~
    & \mathbf{C}^{\mathcal{S}}(\mathbf{x}, \mathbf{q}) = 0 \\
    & \mathbf{C}^{\mathcal{B}}(\mathbf{q}) = 0 \\
    & \mathbf{C}^{\mathcal{D}}(\mathbf{x}) = 0 \\
\end{split}
\end{equation}
where $\mathbf{C}^{\mathcal{S}}$, $\mathbf{C}^{\mathcal{A}}$ and $\mathbf{C}^{\mathcal{D}}$ are the position-based constraints, and $\mathcal{L}$ is the loss function derived from the system states. $\boldsymbol{\lambda}^{t}$ represents the selected gradient variable, which can be control states, system parameters or system states.

\begin{algorithm}[!htbp]
    \caption{Differentiable Framework for Rope-like Objects}
    \label{alg:differentiable_framework_XPBD_rope}
    \SetKwInOut{Input}{Input}
    \SetKwInOut{Output}{Output}


    \tcp{Initialize the gradients variable}
    $\boldsymbol{\lambda}^{t} \leftarrow \lbrace \mathbf{x}^{t}, \mathbf{q}^{t},  \boldsymbol{\eta}_{\mathbf{x}}^\mathcal{S}, \boldsymbol{\eta}_{\mathbf{q}}^\mathcal{S}, \boldsymbol{\eta}_{\mathbf{q}}^\mathcal{B}, \boldsymbol{\eta}_{\mathbf{x}}^\mathcal{D}, \boldsymbol{\eta}_{\mathbf{x}}^\mathcal{G}, \boldsymbol{\eta}^\mathcal{SOR}, \cdots \cdots \rbrace $ \\

        \tcp{Position states Euler prediction}
         $\mathbf{x}^{t+1} \leftarrow \mathbf{x}^{t} + \frac{1}{2} \mathbf{g} \Delta{t}^2 \cdot \boldsymbol{\eta}_{\mathbf{x}}^\mathcal{G}$\\

        
        \tcp{Constraints solving loop}
        \While{iter $<$ iterations}{
            \tcp{Apply shear/stretch constraints using Eq. \ref{eq:shear_stretch_const} }
            $\Delta{\mathbf{x}^{\mathcal{S}}}$, $\Delta{\mathbf{q}^{\mathcal{S}}} \leftarrow \mathtt{solveShearStretch}\left(\mathbf{C}^{\mathcal{S}}\left(\mathbf{x}^{t+1}, \mathbf{q}^{t+1}\right) = 0\right)$\\
            
            \tcp{Apply bend/twist constraints using Eq. \ref{eq:bend_twist_const} }
            $\Delta{\mathbf{q}^{\mathcal{B}}} \leftarrow \mathtt{solveBendTwist}\left(\mathbf{C}^{\mathcal{B}}\left(\mathbf{q}^{t+1}\right) = 0\right)$\\

            \tcp{Apply distance constraints using Eq. \ref{eq:distance_const} }
            $\Delta{\mathbf{x}^{\mathcal{D}}} \leftarrow \mathtt{solveDistance}\left(\mathbf{C}^{\mathcal{D}}\left(\mathbf{x}^{t+1}\right) = 0\right)$ \\
            
            \tcp{Update constraints changes}
            $\mathbf{x}^{t+1} \leftarrow \mathbf{x}^{t+1} + \left(\Delta{\mathbf{x}^{\mathcal{S}}} \cdot \boldsymbol{\eta}_{\mathbf{x}}^\mathcal{S} + \Delta{\mathbf{x}^{\mathcal{D}}} \cdot \boldsymbol{\eta}_{\mathbf{x}}^\mathcal{D} \right)/2 * \boldsymbol{\eta}^\mathcal{SOR}$\\
            $\mathbf{q}^{t+1} \leftarrow \mathbf{q}^{t+1} + \left(\Delta{\mathbf{q}^{\mathcal{S}}} \cdot \boldsymbol{\eta}_{\mathbf{q}}^\mathcal{S} + \Delta{\mathbf{q}^{\mathcal{B}}} \cdot \boldsymbol{\eta}_{\mathbf{q}}^\mathcal{B} \right)/2 * \boldsymbol{\eta}^\mathcal{SOR}$\\
        } 
        

       

        \tcp{Obtain the loss function}
        $\mathcal{L} \leftarrow \mathcal{L} \left( \mathbf{x}^{t+1}, \mathbf{q}^{t+1} \right)$\\

        \tcp{Calculate the gradients}
	    $\displaystyle{\frac{\partial \mathcal{L}}{\partial \boldsymbol{\lambda}^{t}} = \mathtt{autodiff}(\mathcal{L})}$\\
	    
	    \vspace{3mm}
        \Return{$\displaystyle{\frac{\partial \mathcal{L}}{\partial \boldsymbol{\lambda}^{t}} }$}\\
\end{algorithm}

\section{Real-to-Sim and Shape Control Problem Setup}

In this part, we used the proposed differentiable framework to conduct three different experiments both on Rethink Baxter and the da Vinci Research Kit (dVRK) robotic platforms. The Baxter represents a situation of more significant rope manipulation, whereas the dVRK represents a situation of surgical automation involving blood vessel manipulation. We considered the Baxter-rope experiment to involve an inextensible DLO, while the dVRK blood-vessel experiment to be an extensible DLO. 
For both setups, we looked at both a real-to-sim problem (i.e., parameter estimation based on observations from the real world), and a control problem (i.e., using the XPBD model we developed to have rope configuration reach a target configuration iteratively). The implementation of each experiment will be provided later. 

\subsection{Solver Setups}
\textcolor{cmtred}{
In classical PBD solver \cite{Macklin_2017}, the Jacobi approach averages each constraint step changes, and keeps updating iteratively. The convergence compromises between the number of iterations and the satisfaction of other constraints. However, it was not suitable for simulation of the inextensible effects for ropes with significant axial stiffness. 
}

\textcolor{cmtred}{
Precisely, we needed to guarantee the fulfillment of the distance constraint in Eq. \ref{eq:distance_const} completely.
A direct linear solver was proposed \cite{Xu_2018_SurgicalThread} based on the tridiagonal matrix algorithm, i.e., Thomas algorithm, to preserve the inextensible characteristics of rope-like objects. 
We will refer to methods individually as the Jacobi XPBD and Thomas XPBD, respectively.}
\textcolor{cmtred}{All other constraints, such as shear/stretch and bend/twist constraints, used the Jacobi method.\footnote{\textcolor{cmtred}{Thus, in this paper, when Jacobi XPBD and Thomas XPBD are indicated, it is distinguishing how the distance constraint is solved.}}
}

\subsection{Data Preprocessing}

\begin{figure}
\begin{center}
\begin{tabular}{cc}
 \includegraphics[width=0.45\linewidth ]{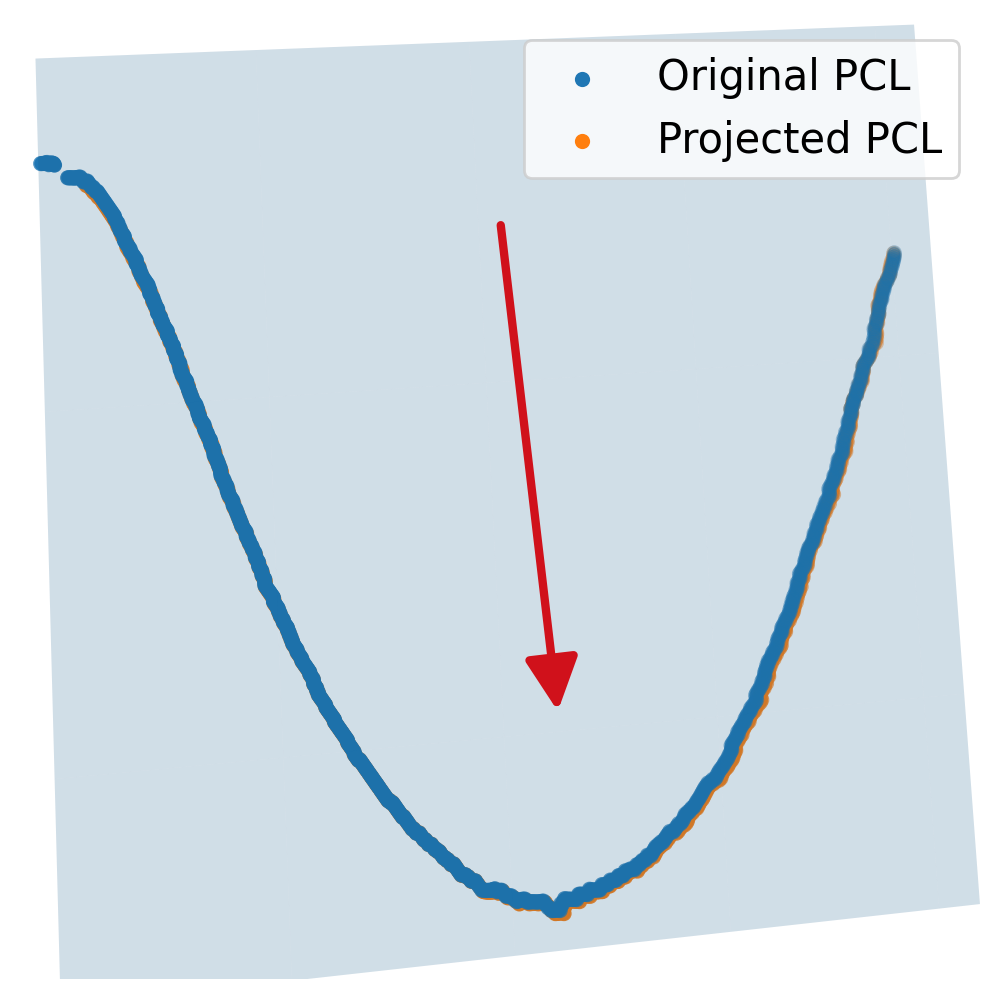} 
 
 & \includegraphics[width=0.45\linewidth]{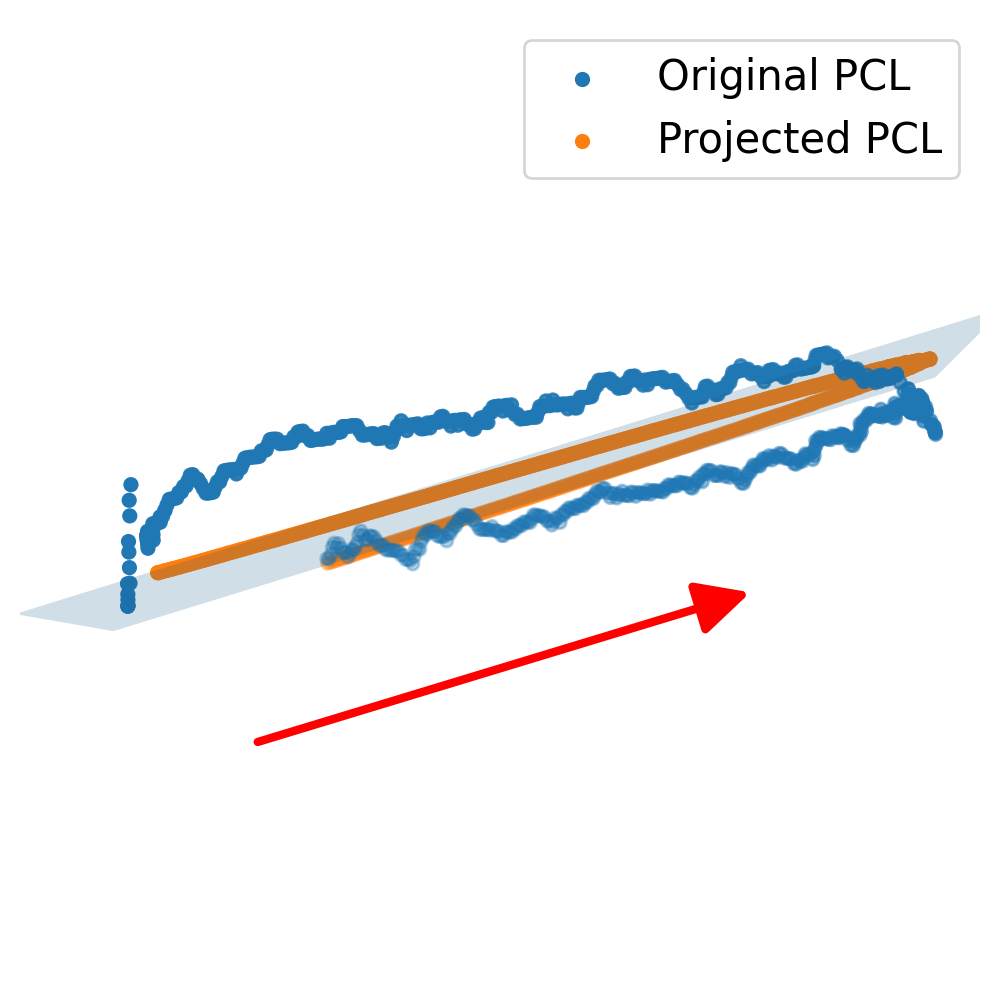} \\
\end{tabular}
\end{center}
\vspace{-0.6cm}
\caption{\textbf{Projecting captured rope point cloud to 2D plane} | Front view (left) and top view (right) for the original point cloud (blue) and projection point cloud (orange). The red arrow is the gravity vector. We projected the point cloud onto the plane defined by the gravity vector and two rope endpoints.}
\label{fig:DATA_PROCESS}
\end{figure}

\textcolor{cmtred}{In our experiments, we obtained both the point cloud and RGB image data using the Microsoft Azure Kinect. For the 2D images obtained, we extracted the centerline of the rope using MATLAB skeletonization function\footnote{https://www.mathworks.com/help/images/ref/bwskel.html}.
}
Since the raw point cloud data was noisy, we projected them to a hyperplane defined by the gravity vector and two endpoints of the rope, as shown in Fig. \ref{fig:DATA_PROCESS}. \textcolor{cmtred}{The projected point cloud was constrained within a plane, which was easier for identification of bend/twist effects.}
\textcolor{cmtred}{We used the de-noised 2D centerline from the image and projected  3D point clouds as ground truth data for loss computation in the following section.}

\subsection{Loss Functions}
\begin{figure}
    \centering
    \includegraphics[width=1.0\linewidth]{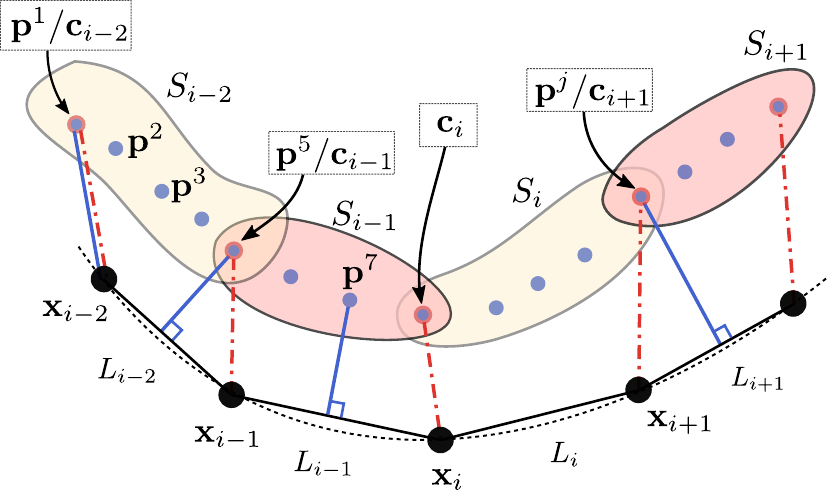}
    \caption{\textbf{Visualization of different position-based losses for shape matching |} The rope was simulated with the particle-based method and represented by connected lines. Curve dotted line was the simulated rope, \textcolor{cmtred}{discretized into particles in black points.} \textcolor{cmtred}{$L_{i}$ is line for neighbouring particles, such as from  particle $\bold{x}_i$ to particle $\bold{x}_{i-1}$
    The preprocessed real data (representing 3D point cloud projected along gravity or 2D centerline) were split into different segments (i.e. $S_{i-2}$, $S_{i-1}$, etc). Points of real data $\mathbf{p}^j$ are in blue, and the start/endpoints (i.e., $\mathbf{c}_{i-2}$, $\mathbf{c}_{i-1}$)  of each segment group set are in blue and circled in red. The solid blue lines indicate the minimum distance from the point cloud to the simulated rope particles. The dashed-point red lines indicate correspondence of each segment's start/endpoints to the simulated rope particles.}
    }
    \label{fig:loss_projection}
\end{figure}

We defined four types of primary losses to investigate for shape matching. We looked at the combinations of these losses for ablation, resulting in 9 different aggregate loss functions for optimization. The loss functions are shown in Table \ref{tab:nine_experiment_losses}.

 \begin{itemize}
    \item \textcolor{cmtred}{\textit{Point-to-Line} : As shown in Fig. \ref{fig:loss_projection}, we intended to find the minimum distance from each preprocessed real data point $\mathbf{p}_j$,
    to the line segments $L_i$ between each pair of adjacent simulated particle $\left(\mathbf{x}_i, \mathbf{x}_{i+1}\right)$. Thus, the primary loss regarding point-to-line loss was
    }
    \begin{equation}
     \textcolor{cmtred}{
        \mathcal{L}_{PL}= \sum_{j=1}^{N}  \min_{ \tilde{\mathbf{p}}  \in
        \bigcup\limits_{k=1}^{K-1} L_{k}}
        \norm{\mathbf{p}^j - \tilde{\mathbf{p}} }
     }
    \end{equation}
    
    \textcolor{cmtred}{where we discretized the rope to $K$ particle nodes, along $N$ preprocessed real data points, and $\tilde{\mathbf{p}}$ was the projected locations on each line segment with minimum distance.}
    
    \item \textcolor{cmtred}{\textit{Segment-to-Particle} : We splitted the real data points (preprocessed for noises and re-ordered) into $K$ different segments along the vector formulated by two rope endpoints, i.e. $S_{i-2}$, $S_{i-1}$ etc shown in Fig. \ref{fig:loss_projection}. The start/endpoints (i.e., $\mathbf{c}_{i-2}$, $\mathbf{c}_{i-1}$) of each segment were determined according to the proportional length to the whole data set. We evaluated the distance between the start/endpoints of each segment and the corresponding particle in the simulation. Then, the segment-to-particle loss was,
     }
     \begin{equation}
     \textcolor{cmtred}{
        \mathcal{L}_{SP} =  \sum_{i=1}^{K} \norm{\mathbf{c}_i - \mathbf{x}_i}
     }
    \end{equation}

    \item \textcolor{cmtred}{\textit{Segment-to-Line} : As above, we already divided the real data into $K$ segments. The projection distance between all points in each segment set ($S_i$) and the corresponding line segment ($L_i$) could be used to evaluate the loss. Thus, the segment-to-line loss was,
    }
    \begin{equation}
     \textcolor{cmtred}{
        \mathcal{L}_{SL}= \sum_{i=1}^{K} \sum_{ \mathbf{p}^j \in S_i}^{M_k}  \min_{\tilde{\mathbf{p}} \in L_{k}}
        \norm{\mathbf{p}^j - \tilde{\mathbf{p}}}
     }
    \end{equation}
    \textcolor{cmtred}{where $M_k$ was the number of real data points for each segment group set, and $\tilde{\mathbf{p}}$ was the projected locations on the line segment with minimum distance.}
    
    \item \textcolor{cmtred}{\textit{Lowest 3D Point Along Gravity Vector} : The lowest point along the gravity direction of the denoised real data was compared with the one in simulation. This could compensate for lacking depth information while using the 2D centerline. The loss was defined as,
    }
    \textcolor{cmtred}{
    \begin{equation}
    \begin{split}
        \mathcal{L}_{LO} = \norm{\mathbf{p}^{lowest} - \left(\bigcup\limits_{k=1}^{K-1} L_{k}\right)^{lowest} }
    \end{split}
    \end{equation}
    }
\end{itemize}

\textcolor{cmtred}{In our experiments, the point-to-line, segment-to-particle, and segment-to-line losses could be applied to either the 3D point cloud or 2D centerline. We performed 9 different combinations of types of losses as shown in Table \ref{tab:nine_experiment_losses}. The loss function $\mathtt{OBJ1}$, $\mathtt{OBJ4}$, $\mathtt{OBJ7}$ contained 3D information only, while $\mathtt{OBJ2}$, $\mathtt{OBJ5}$, $\mathtt{OBJ8}$ were considering 2D cases only. $\mathtt{OBJ3}$, $\mathtt{OBJ6}$, $\mathtt{OBJ9}$ were applied to both 2D cases and the included 3D lowest point with depth information.
}

\begin{table*}
\renewcommand{\arraystretch}{1.2}
\begin{center}
\begin{tabular}{ | c | c | c | c | c | c | c | c | c | c | c |}
 \hline
 \textbf{Symbol} & $\mathtt{OBJ1}$ & $\mathtt{OBJ2}$ & $\mathtt{OBJ3}$ & $\mathtt{OBJ4}$ & $\mathtt{OBJ5}$ & $\mathtt{OBJ6}$ & $\mathtt{OBJ7}$ & $\mathtt{OBJ8}$    & $\mathtt{OBJ9}$ \\
 \hline
 \textbf{Loss} & $\mathcal{L}^{3D}_{PL}$ & $\mathcal{L}^{2D}_{PL}$ & $\mathcal{L}^{2D}_{PL}+\mathcal{L}_{LO}$ & $\mathcal{L}^{3D}_{SP}$ &  $\mathcal{L}^{2D}_{SP}$  &  $\mathcal{L}^{2D}_{SP}+\mathcal{L}_{LO}$  & $\mathcal{L}^{3D}_{SL}$ &  $\mathcal{L}^{2D}_{SL}$  &  $\mathcal{L}^{2D}_{SL}+\mathcal{L}_{LO}$  \\
 \hline
 \textbf{Meaning} & \multicolumn{3}{c|}{Point-to-Line for 3D and 2D}  & \multicolumn{3}{c|}{Segment-to-Particle for 3D and 2D} & \multicolumn{3}{c|}{Segment-to-Line for 3D and 2D} \\
 \hline
\end{tabular}
\end{center}
\caption{\textbf{Loss type} | We used nine types of loss functions for our differential framework and made a comparison over these loss functions. 
}
\label{tab:nine_experiment_losses}
\end{table*}

\section{Experiments And Result Analysis}

\subsection{Inextensible Rope Parameter Identification}
\label{sec:inextensible_para_identi_result}
Parameter identification and \textcolor{cmtorange}{position estimation of} the control point were carried out on an inextensible rope manipulated by the Baxter robot. \textcolor{cmtorange}{We fixed one endpoint of the rope to the environment, and the other endpoint of the rope to the manipulator end-effector (referred as the control point). We \textcolor{cmtgreen}{continuously} moved the control point and collected 36 frames of point cloud for the rope deformation. \textcolor{cmtgreen}{The Microsoft Azure Kinect was used to obtain the 2D RGB image frames for centerline extraction at the same time. Since only translational movement (fixed rotation angle) of end-effector is controlled}, the point cloud had occlusions and \textcolor{cmtgreen}{was not fully visible at the local region} near the control point. \textcolor{cmtgreen}{We applied the keypoint-based kinematics reconstruction from \cite{lu2021pose} to identify the 3D position of the control point. Therefore, the ground truth data for 3D point cloud and 2D centerline, as well as control point were ready for real-to-sim transfer.}
} 

\begin{figure}
\begin{center}
\begin{tabular}{cc}
\includegraphics[width=0.45\linewidth]{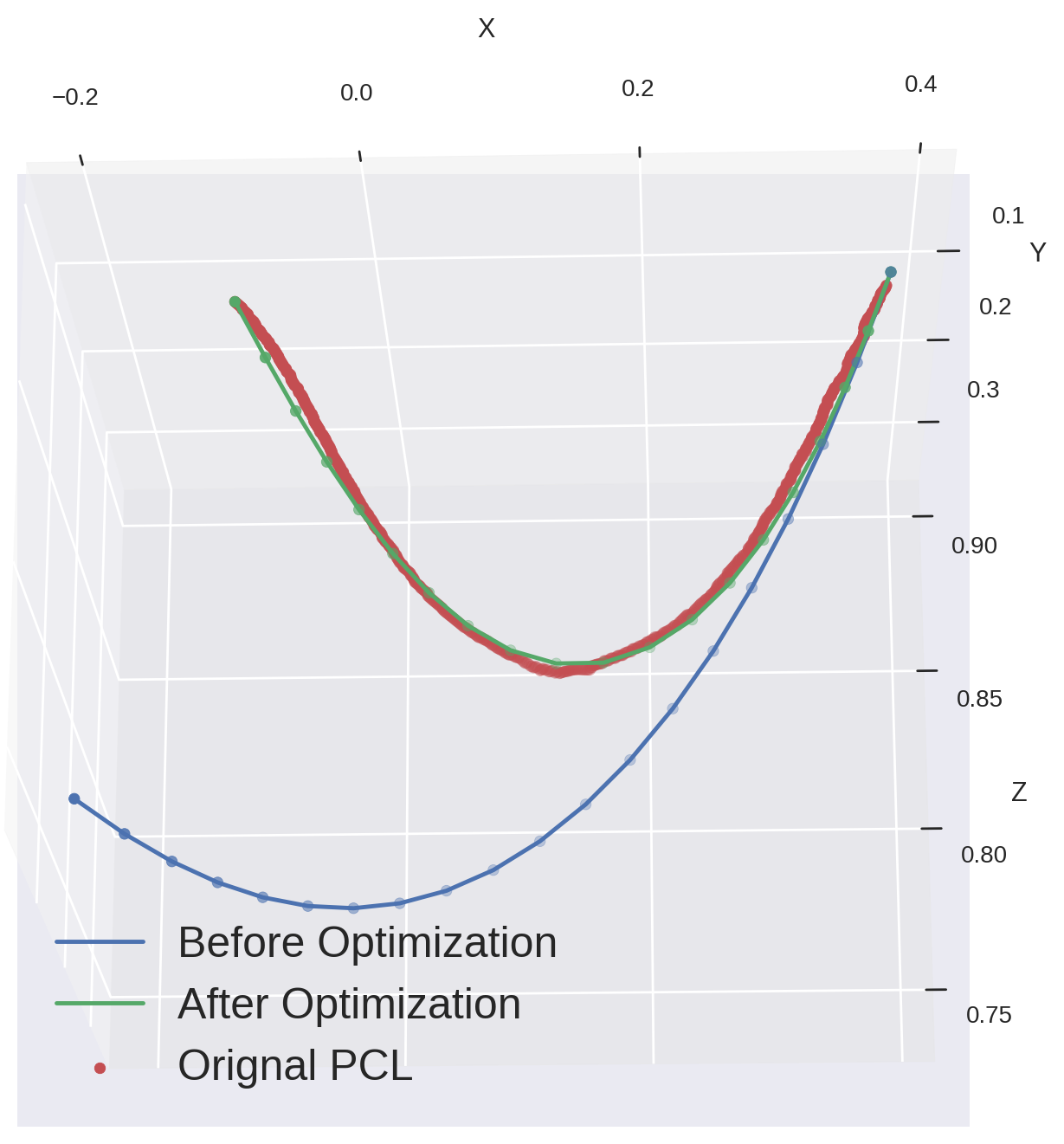}
&
\includegraphics[width=0.45\linewidth]{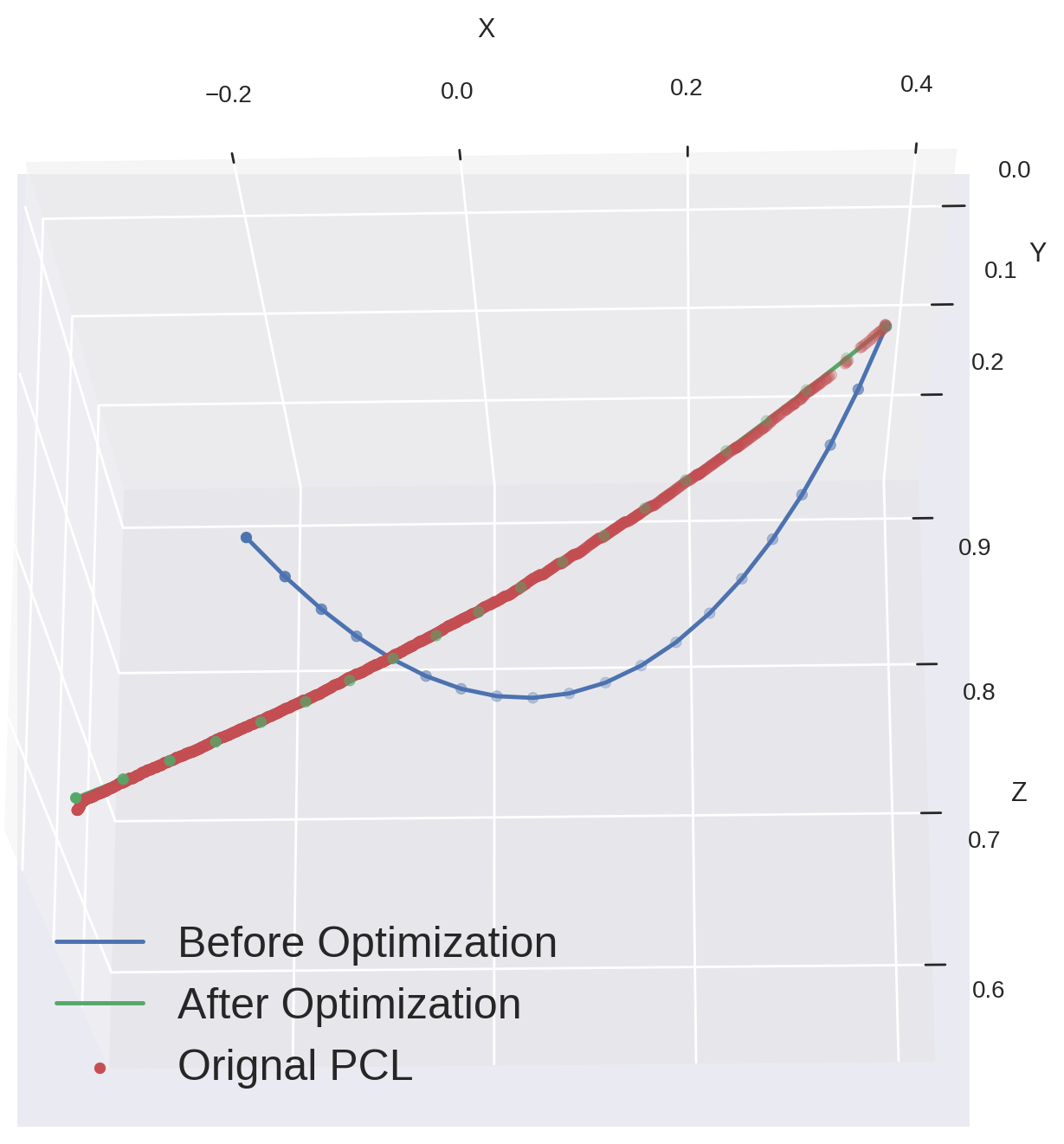}
\end{tabular}
\end{center}
\vspace{-0.6cm}
\caption{\textbf{Parameter optimization results for inextensible rope} (i.e., real-to-sim result) | The bending (left) and the tensioned (right) status of the \textcolor{cmtred}{simulated} rope before and after parameter optimization. PCL, in red, is the real point cloud that the simulation is trying to match up to.}
\label{fig:ParameterOptimization}
\end{figure}

\textcolor{cmtorange}{\textcolor{cmtgreen}{In XPBD simulation}, considering the computational cost and stiffness of the rope, the number of particle were set by $K=20$. $\mathbf{x}_{0}$ was the control point and $\mathbf{x}_{19}$ corresponding to the endpoint \textcolor{cmtgreen}{was fixed} to the environment.}
\textcolor{cmtorange}{According to Section \ref{sec:real2sim_parameters}, there were six constraint stiffness parameters needed to estimate, as shown in Table \ref{tab:Parameter}. Because 3D point cloud could be more informative than 2D centerline \textcolor{cmtgreen}{which lacked depth data}, the parameters were optimized with $\mathtt{OBJ1}$ \textcolor{cmtgreen}{using the 3D point-to-line loss}. We then estimated the position of the control point (i.e. $\boldsymbol{\lambda}=\mathbf{x}_{0}$, \textcolor{cmtgreen}{see Algorithm \ref{alg:differentiable_framework_XPBD_rope}}) using all loss functions in Tab. \ref{tab:nine_experiment_losses} to evaluate the accuracy of parameter identification.}
\textcolor{cmtorange}{For parameter identification,} \textcolor{cmtgreen}{a single gravity parameter was inferred} based on Thomas solver, the comparison of simulated results before and after parameter optimization was shown in Fig. \ref{fig:ParameterOptimization}.  \textcolor{cmtred}{After parameter optimization, the simulation states was approximately approaching to the ground truth point cloud.} As for the Jacobi solver, it needed two \textcolor{cmtgreen}{sets} of gravity stiffness to deal with \textcolor{cmtred}{different convergency of distance constraint} for \textcolor{cmtgreen}{varied} rope states (bending and tensioned), \textcolor{cmtorange}{as shown in Table. \ref{tab:Parameter_Jacobi}.}
\begin{table}[]
\renewcommand{\arraystretch}{1.2}
\begin{center}
\begin{tabular}{ c c c c}
 \hline
 Parameter & Initial & Optimized & Meaning \\ 
 \hline
 $\boldsymbol{\eta}_{\mathbf{x}}^\mathcal{G}$ & 0.04 & 0.024 & position stiffness (gravity)\\  
 $\boldsymbol{\eta}_{\mathbf{x}}^\mathcal{D}$ & 1.0 & 0.48  & position stiffness (distance)\\  
 $\boldsymbol{\eta}_{\mathbf{x}}^\mathcal{S}$ & 1.0 & 1.16  & position stiffness (shear/stretch)\\  
 $\boldsymbol{\eta}_{\mathbf{q}}^\mathcal{S}$ & 1.0 & 0.52  & quaternion  stiffness (shear/stretch)\\  
 $\boldsymbol{\eta}_{\mathbf{q}}^\mathcal{B}$ & 1.0 & 1.49  & quaternion stiffness (bending/twist)\\  
 $\boldsymbol{\eta}^\mathcal{SOR}$ & 1.0 & 0.61  & successive over-relaxation weight \\ 
  \hline
\end{tabular}
\end{center}

\caption{\label{tab:Parameter} \textcolor{cmtorange}{\textbf{Parameter estimation result for inextensible rope}} | \textcolor{cmtred}{The parameter estimation result of the differentiable Thomas XPBD solver for the inextensible rope.}
}
\end{table}

\begin{figure}
\vspace{2mm}
\centering
\includegraphics[width=\linewidth]{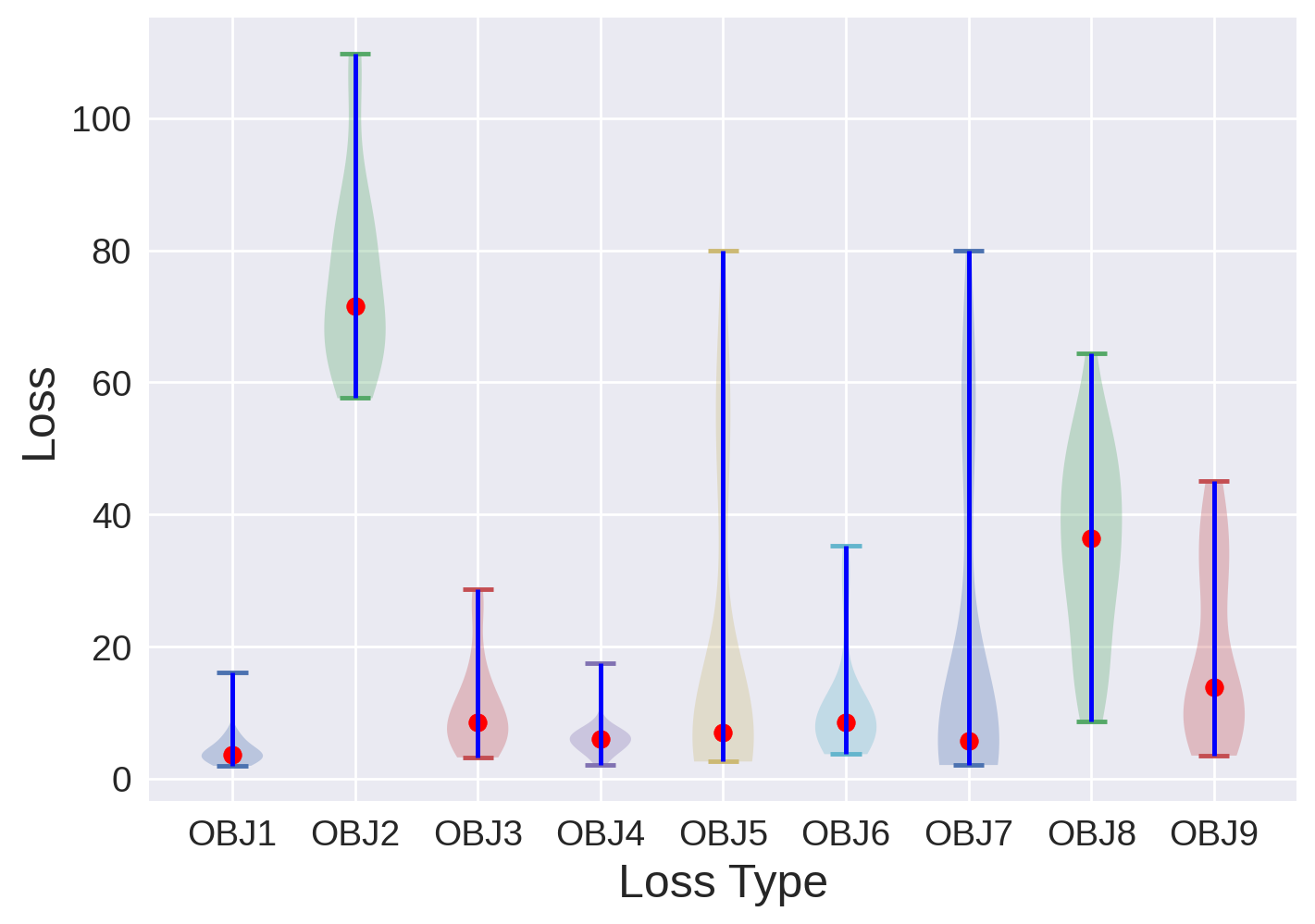}

\caption{\textcolor{cmtorange}{\textbf{Losses on inextensible rope as a function of objective} } | \
\textcolor{cmtorange}{ By using different loss functions\ref{tab:nine_experiment_losses}, we obtained nine kinds of simulation results for all frames. We used OBJ1 to evaluate the deviation caused by different loss functions.  }
The simulation result got from $\mathtt{OBJ1}$ has the smallest deviation compared with the ground truth one.
}
\label{fig:Loss9}
\vspace{-0.2in}
\end{figure}

\begin{figure*}[htbp]
\vspace{2mm}
\centering
\includegraphics[width=\linewidth]{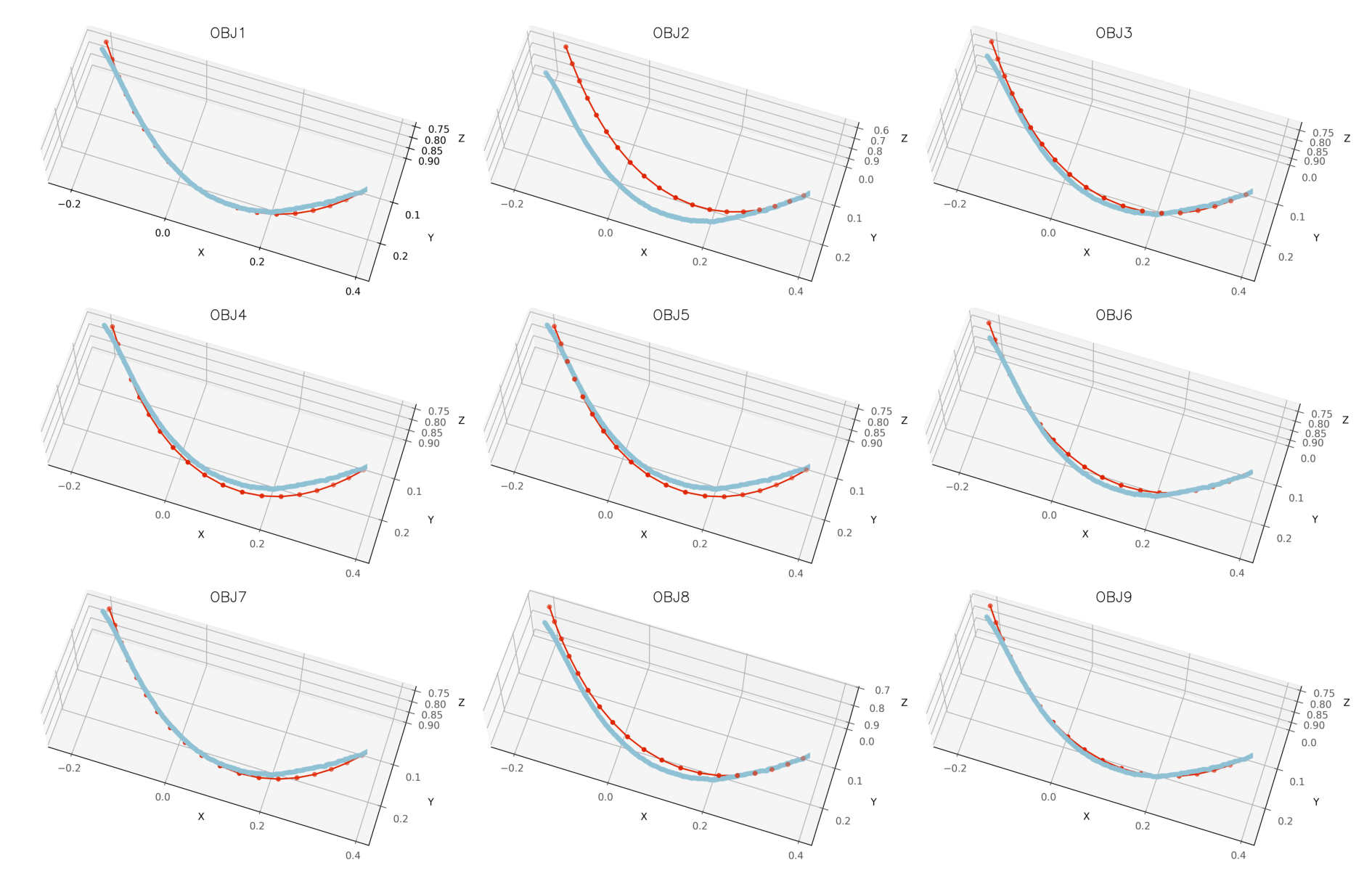}
\vspace{-0.6cm}

\caption{\textcolor{cmtorange}{\textbf{Deformation estimation for inextensible rope using different loss functions | }}
From \textbf{Right} to \textbf{Left } and from \textbf{Top} to \textbf{Bottom}: The simulation result for 9 different loss functions based on Thomas XPBD solver. Blue lines represented ground truth provided by the point clouds. Red lines represented the optimized result regarding different loss functions.
}
\label{fig:PlotDifferenLossSimulatedResult}
\end{figure*}

\textcolor{red}{}
\textcolor{cmtred}{\textcolor{cmtorange}{For position estimation of control point, we made a comparison over the nine kinds of loss functions shown in Table \ref{tab:nine_experiment_losses} firstly}. The 3D point cloud included the rope information in 3D space, while the 2D centerline did not contain depth information. As a result, the simulation result containing the 3D information showed the best performance, i.e., $\mathtt{OBJ1}$, $\mathtt{OBJ4}$, and $\mathtt{OBJ7}$ as shown in the Fig. \ref{fig:PlotDifferenLossSimulatedResult}. For $\mathtt{OBJ3}$, $\mathtt{OBJ6}$, and $\mathtt{OBJ9}$, whose loss function consisted of 2D information and partial 3D information (the lowest point along gravity direction), the simulation result was improved compared to $\mathtt{OBJ2}$, $\mathtt{OBJ5}$, and $\mathtt{OBJ8}$, which only considered 2D information. As shown in Fig.  \ref{fig:PlotDifferenLossSimulatedResult} and Fig. \ref{fig:Loss9} , $\mathtt{OBJ1}$ obtained the best performance among all these losses. \textcolor{cmtgreen}{It simply proved the accuracy of parameter identification, and showed that the 3D data contributed more than 2D cases for control point estimation.}
}

\begin{figure*}[htbp]
\centering
\includegraphics[width=\linewidth]{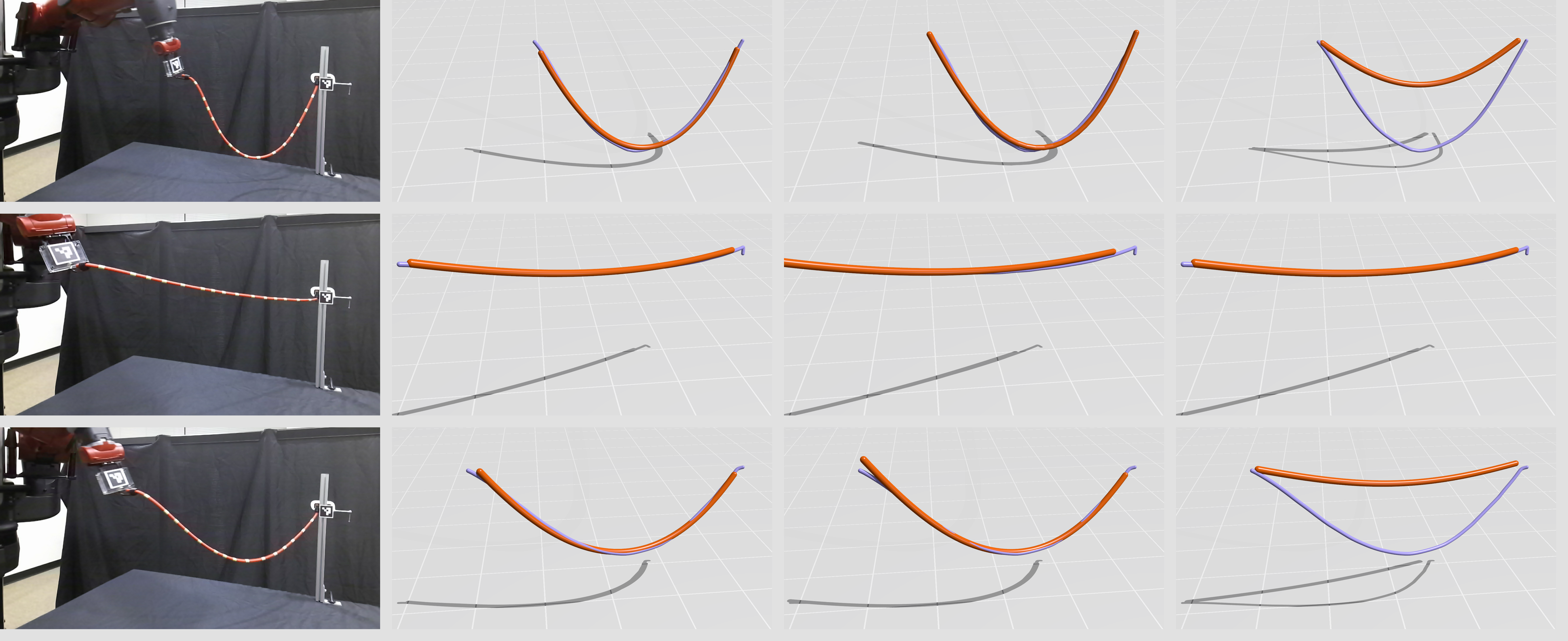}
\caption{
\textcolor{cmtgreen}{\textbf{Deformation estimation for inextensible rope on Baxter} |}
\textcolor{cmtred}{
\textbf{Each Column}: (1) Raw RGB image. (2) The simulation result based on Thomas XPBD, with good convergency for both bending and tensioned status. The simulation result with gravity stiffness of (3) $\boldsymbol{\eta}_{\mathbf{x}}^\mathcal{G}=0.26$ (better convergency for bending status and \textcolor{cmtorange}{the length of tensioned rope in the middle exceeds the normal length}) and (4) $\boldsymbol{\eta}_{\mathbf{x}}^\mathcal{G}=0.001$ (better convergency for tensioned status \textcolor{cmtorange}{and the elasticity of the bending ropes in the top and bottom fail to reach the expected.}) from Jacobi XPBD solver. Blue ropes represented ground truth provided by the point clouds. Orange ropes represented the optimized deformation of the rope.
} \textcolor{cmtorange}{ For the light occlusion, we used keypoint detection\cite{lu2021pose} and point cloud instead of Aruco Marker for the location identification of end-effector and fixed point respectively. }
}
\label{fig:Connect_XPBD_Thomas}
\end{figure*}

\begin{figure}

\includegraphics[width=0.5\textwidth]{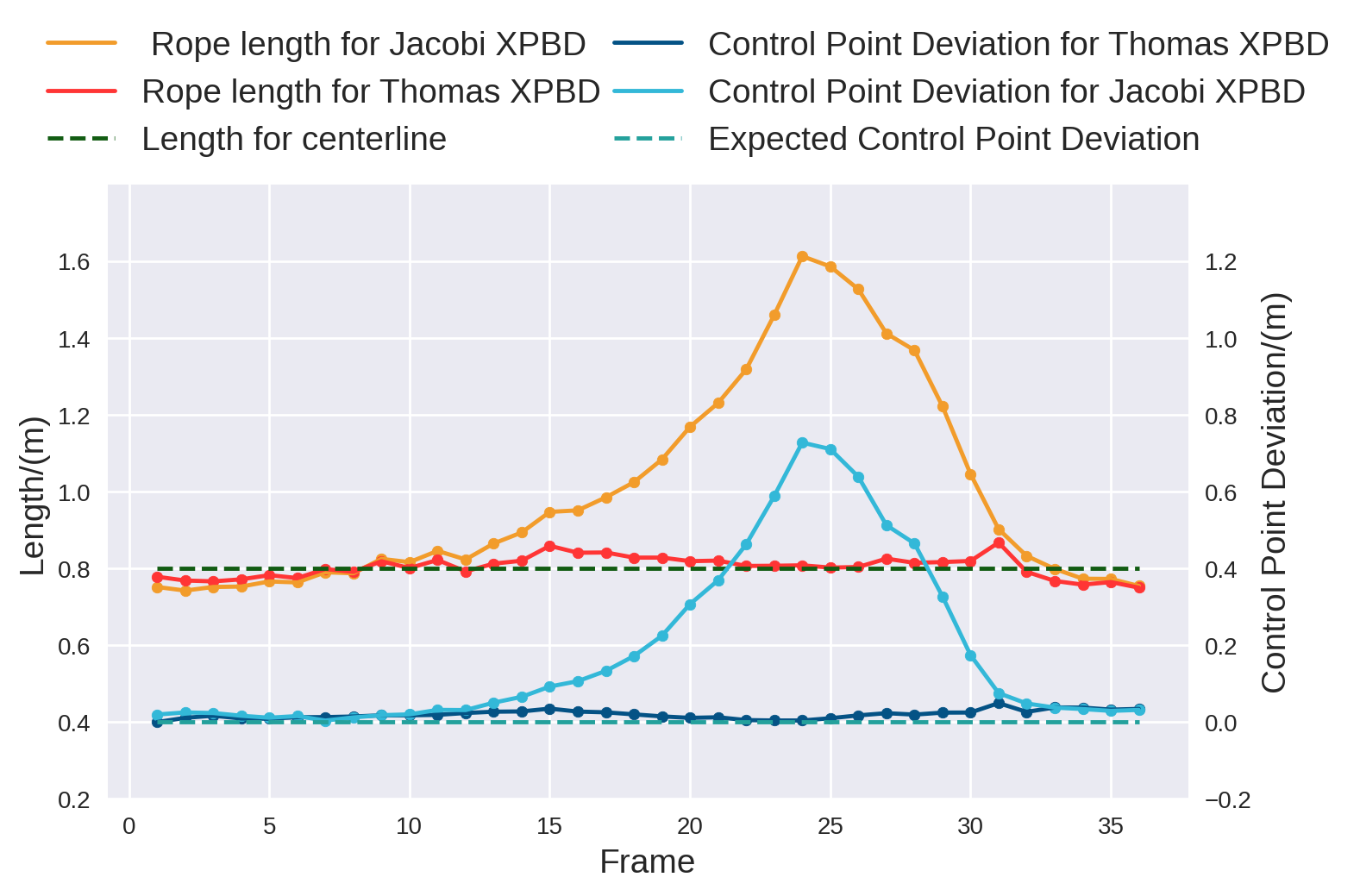}
\vspace{-0.6cm}
\caption{\textbf{Comparison of Thomas and Jacobi XPBD solver on inextensible rope} |  Control point deviation and  rope length regarding different frames of simulation using Thomas XPBD and Jacobi XPBD.
}
\label{fig:Compare}
\end{figure}

\begin{figure*}[htbp]
\vspace{2mm}
\centering
\includegraphics[width=1.0\linewidth]{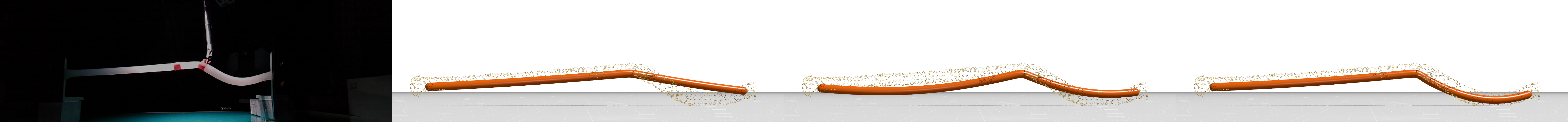}

\caption{ \textcolor{cmtorange}{\textbf{Parameter identification for extensible rope on dVRK} | This experiment shows the simulation result with different $\boldsymbol{\eta}_{\mathbf{x}}^\mathcal{G}$. From left to right: (1) Original RGB image from Azure Kinect. (2) Both sides were $\boldsymbol{\eta}_{\mathbf{x}}^\mathcal{G}=3$. (3) Both sides were $\boldsymbol{\eta}_{\mathbf{x}}^\mathcal{G}=25$. (4) The relax side was $\boldsymbol{\eta}_{\mathbf{x}}^\mathcal{G}=25$ and the extensible (tensioned) side was $\boldsymbol{\eta}_{\mathbf{x}}^\mathcal{G}=3$. The simulation result got from (4) was the closest to the ground truth.}
}
\label{fig:ConnectImageDifferentWeight}
\end{figure*}

\begin{figure*}[htbp]
\centering
\includegraphics[width=1.0\linewidth]{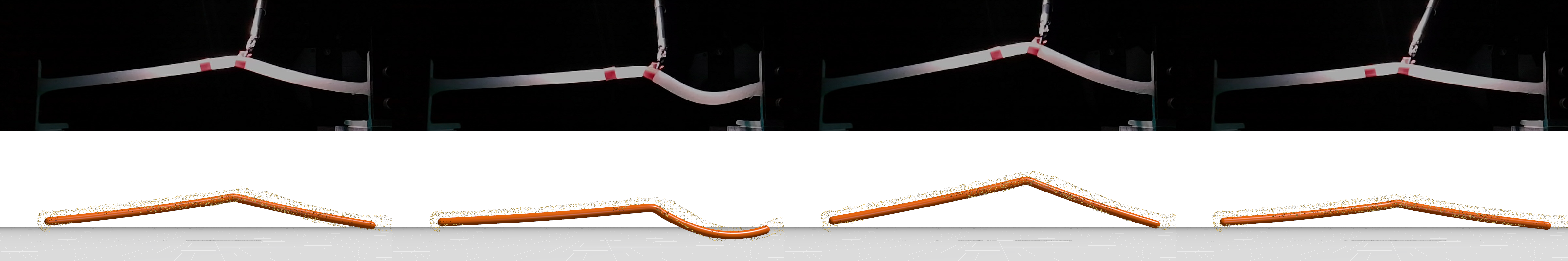}

\caption{\textcolor{cmtgreen}{\textbf{Deformation identification and key points estimation for extensible rope on dVRK} |} \textbf{Top}: Original RGB image from Azure Kinect. \textbf{bottom}: Simulation result for the extensible rope by using Jacobi XPBD solver. Orange ropes represented the optimized deformation of the rope. The manipulation policy from the  differential framework made the simulated result approaches the ground truth one. \textcolor{cmtorange}{The right red marker is the control point and the left one is the reference marker}. }
\label{fig:ConnectImage11}
\end{figure*}

\begin{figure*}
\begin{center}
\begin{tabular}{c c}

     \includegraphics[width=0.4\textwidth]{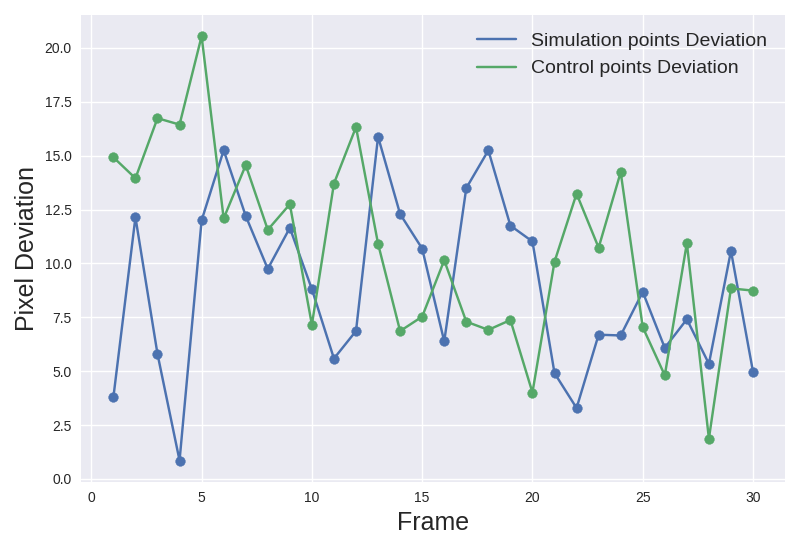} &
     \includegraphics[width=0.4\textwidth]{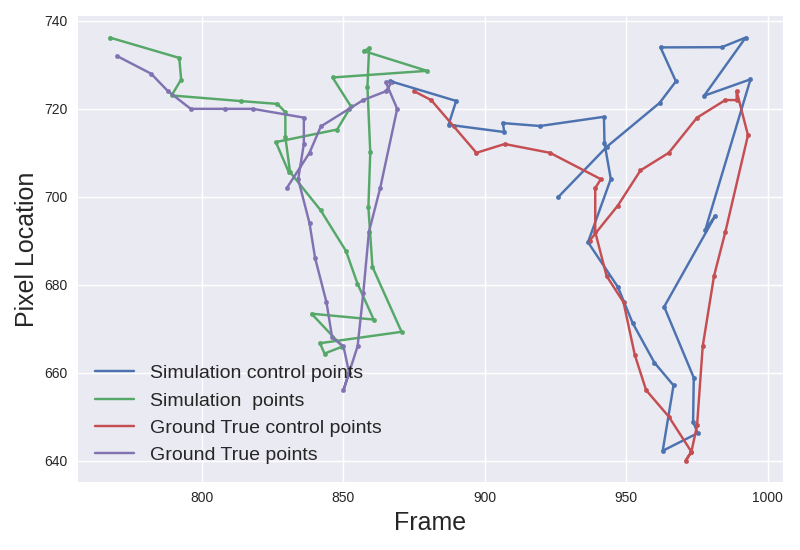} \\
\end{tabular}
\end{center}
\caption{\textcolor{cmtgreen}{\textbf{Trajectory and Deviation of control/reference points for extensible rope} |}
\textbf{Left}: \textcolor{cmtorange}{The pixel deviation of the control point and a reference point in 2D image regarding frames.} \textbf{Right}: Control point and reference points trajectory for the real case and the simulation one.}
\label{fig:DVRK_reprojection}
\end{figure*}



\begin{figure*}[htbp]
\begin{center}
\begin{tabular}{ccc}
\vspace{-0.2cm}
 \includegraphics[width=0.32\textwidth]{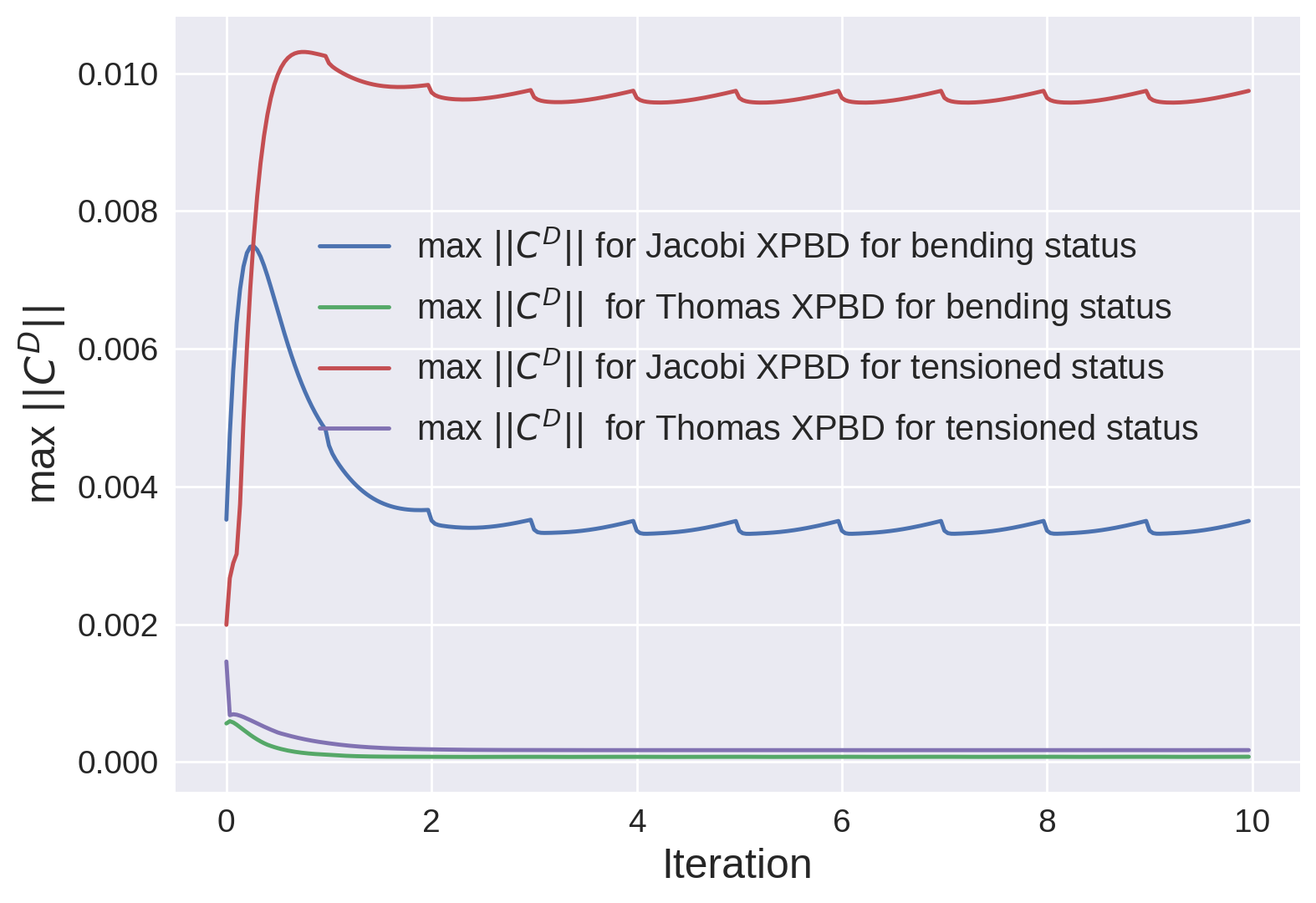} 
 & \includegraphics[width=0.32\textwidth]{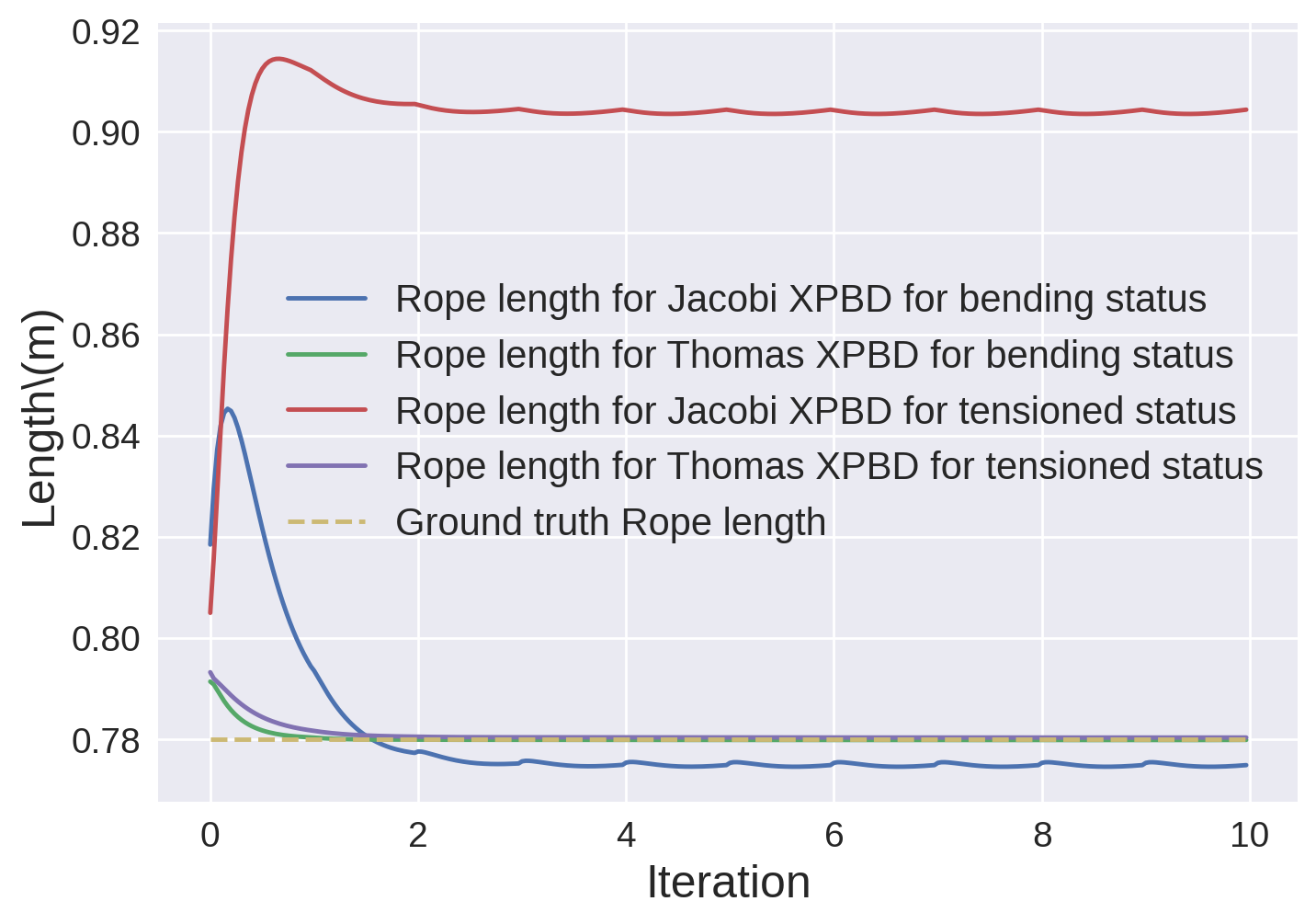} & \includegraphics[width=0.32\textwidth]{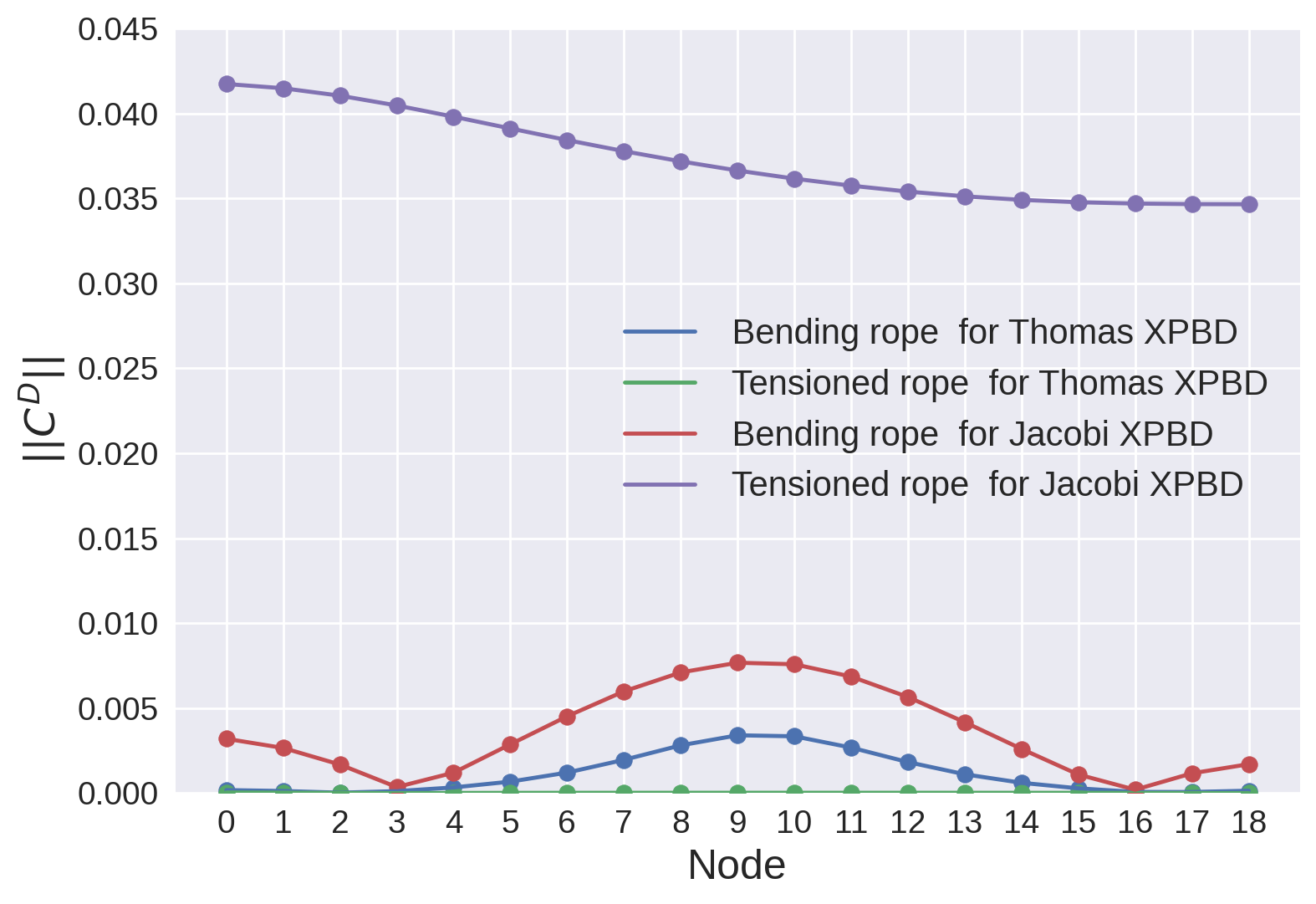} \\
\end{tabular}
\end{center}
\vspace{-0.6cm}
\caption{\textcolor{cmtorange}{\textbf{Simulation result analysis for \textcolor{cmtgreen}{distance constraint} of inextensible rope on Baxter}} | \textcolor{cmtred}{(1) The evolution  of maximal distance constraint value of all nodes (i.e., \textcolor{cmtred}{$\mathrm{max} \norm{ \mathbf{C}^{\mathcal{D}} } $}) regarding iterations. (2) The evolution of rope length regarding iterations. (3) The $\norm{\mathbf{C}^{\mathcal{D}}}$ value of each pair of neighbouring nodes of the last iteration. Both the tensioned and bending status are shown based on differentiable Thomas solver and Jacobi solver of XPBD simulation.\textcolor{cmtorange}{( $\boldsymbol{\eta}_{\mathbf{x}}^\mathcal{G}=0.26$  for Jacobi XPBD solver.)}
} 
}
\label{fig:Thomas_XPBD}
\end{figure*}

\textcolor{cmtorange}{We also compared the performance of Thomas XPBD and Jacobi XPBD when estimating the position of the control point.} \textcolor{cmtgreen}{The middle plot} in Fig. \ref{fig:Thomas_XPBD} showed the rope length change of the simulated result from 36 frames. The average length deviation over the ground truth length for Thomas XPBD solver was $2\% $ for all structures, but for Jacobi solver, it was $26.6\%$. Especially when the rope became tense, the error for Jacobi solver became more significant. The same phenomenon occurred on the error of the control point's position in Fig. \ref{fig:Compare}. \textcolor{cmtgreen}{The left and right plots in} Fig. \ref{fig:Thomas_XPBD} indicated that the distance constraint in the Jacobi XPBD was hard to be satisfied when the rope was approaching the tensioned status.
\textcolor{cmtgreen}{The Jacobi XPBD intuitively tried to comprise the satisfaction of each constraint solving. It iteratively struggled to maintain the stretch, twist, and distance against gravity. Thus, it resulted in a no-complete convergency of distance constraint $\mathbf{C}^{\mathcal{D}}$.}
For Thomas XPBD, the distance constraint would be satisfied between each neighboring node simultaneously \textcolor{cmtgreen}{within one step solving}, which was \textcolor{cmtgreen}{similar} as the position update mechanism of the gravity. In this case, Thomas XPBD solver ensured that the length of rope was unchanged, and it was more suitable for the simulation of inextensible rope. \textcolor{cmtorange}{The simulation result shown in Fig. \ref{fig:Connect_XPBD_Thomas} proved our conclusion.}

\subsection{\textcolor{cmtgreen}{Extensible Rope Parameter Identification and Key Points Estimation}} 
Parameter identification and \textcolor{cmtorange}{position estimation of} control point were carried out on an extensible, flexible silicone \textcolor{cmtgreen}{rod} using the dVRK surgical robot. The silcone rod was chosen to resemble a compliant and dissected vessel, a technique that is often used in surgery to avoid damaging the vessel. The two endpoints of the rope were fixed, and the control point was in the middle of the rope. \textcolor{cmtorange}{We moved the control point \textcolor{cmtgreen}{to collect} 30 frames of point cloud data.} 
\textcolor{cmtorange}{In simulation, the number of \textcolor{cmtgreen}{particle} nodes were $K=40$.} \textcolor{cmtgreen}{$\bold{x}_{0}$ and $\bold{x}_{39}$ corresponding to both the endpoints were fixed and $\bold{x}_{20}$ was the control point. We did not need to preserve the inextensible behavior completely since the silicon rope had a lower axial stiffness.} Thus, we only implemented the Jacobi XPBD solver for the distance constraint in this experiment.
\textcolor{cmtorange}{The loss function used was $\mathtt{OBJ1}$ in Table \ref{tab:nine_experiment_losses}, and we estimated the same parameters as the ones indicated in Section \ref{sec:real2sim_parameters}. After parameter estimation, we estimated the position of the control point (i.e. $\boldsymbol{\lambda}=\bold{x}_{19}$) using $\mathtt{OBJ1}$ and $\mathtt{OBJ2}$, \textcolor{cmtgreen}{i.e., point-to-line for both 3D and 2D cases.}
}
\textcolor{cmtorange}{For parameter identification,}
we used grid-search to identify the optimized six parameters in \textcolor{cmtorange}{Table \ref{tab:Parameter_Jacobi}} for extensible rope. Since the control point was set at the middle of the rope, manipulation could result in one side being tight and the other being loose. Thus, the effect of the gravity over two sides was different, and we needed to define different gravity weights $\boldsymbol{\eta}_{\mathbf{x}}^\mathcal{G}$ for the two sides, as shown in Fig. \ref{fig:ConnectImageDifferentWeight}. 

\begin{table}[]
\renewcommand{\arraystretch}{1.2}
\begin{center}
\begin{tabular}{ c c  c}
 \hline
 Parameter  & Inextensible rope & extensible rope \\ 
 \hline
 $\boldsymbol{\eta}_{\mathbf{x}}^\mathcal{G}$ & 0.26 (bending) / 0.001 (tensioned) & 25 (relax) / 3 (tensioned)\\  
 $\boldsymbol{\eta}_{\mathbf{x}}^\mathcal{D}$ &1.19 & 0.87  \\  
 $\boldsymbol{\eta}_{\mathbf{x}}^\mathcal{S}$ & 0.80 & 0.873  \\  
 $\boldsymbol{\eta}_{\mathbf{q}}^\mathcal{S}$ & 1.19 & 1.0\\  
 $\boldsymbol{\eta}_{\mathbf{q}}^\mathcal{B}$ & 0.80 & 1.30  \\  
 $\boldsymbol{\eta}^\mathcal{SOR}$ & 0.79 & 0.61  \\ 
  \hline
\end{tabular}
\end{center}

\caption{ \textcolor{cmtorange}{\textbf{Parameter estimation result for Jacobi solver on inextensible and extensible rope}} | \textcolor{cmtred}{The parameter estimation result of the differentiable \textcolor{cmtorange}{Jacobi XPBD solver for the inextensible  and extensible rope.}}
}
\label{tab:Parameter_Jacobi}
\end{table}

\textcolor{cmtorange}{With the optimized parameters, we estimated the position of the control point and a reference point, \textcolor{cmtgreen}{namely by key points.}}
Apart from the control point, there was a reference point with a red marker, \textcolor{cmtgreen}{shown in Fig. \ref{fig:ConnectImageDifferentWeight}}. The marker was used only for comparison between ground truth trajectory and simulation result in Fig. \ref{fig:DVRK_reprojection} \textbf{Right} and not for parameter estimation.
Fig. \ref{fig:DVRK_reprojection} 
\textbf{Left} showed the error curve of the location of reference and control point. Even though the raw data of the extensible rope was noisier, the average error of the control point and the reference point was acceptable, which proved the robustness of our solver.
In Fig. \ref{fig:ConnectImage11}, we showed the original image and simulation result for different frames, and no matter how the rope deformed, our solver could get a desirable outcome. Even though Thomas solver outperformed the Jacobi solver in the previous experiment, Thomas XPBD was too tough to simulate the silicon rope. Thomas XPBD required the length of the rope to remain unchanged, which contradicted the silicon rope's \textcolor{cmtred}{extensible} property. Thus, the Thomas solver diverged for the simulation, and the Jacobi solver was more suitable.

\begin{figure*}
\begin{centering}
\includegraphics[width=\linewidth]{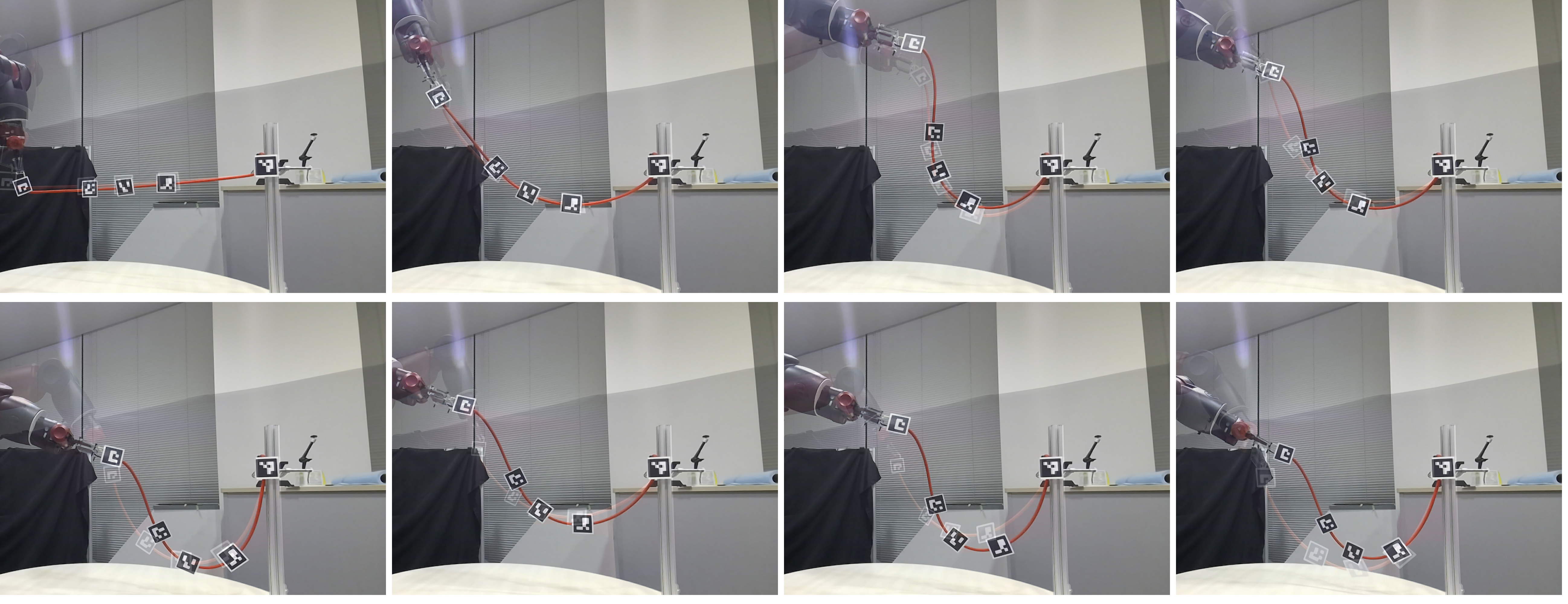}
\end{centering}

\caption{\textcolor{cmtgreen}{\textbf{Shape control for inextensible rope on Baxter} | \textbf{Top Row}: The ideal shape control task by identified parameters. \textbf{Bottom Row}: The not-ideal cases. The target control shape for ropes are in low opacity, while our control results are in solid opacity.} }
\label{fig:real2sim}
\end{figure*}

\subsection{\textcolor{cmtred}{Inextensible Rope Shape Control}}

\textcolor{cmtorange}{Based on the Thomas XPBD solver, the shape control task was carried out on the same Baxter setup as the Section \ref{sec:inextensible_para_identi_result}.}
\textcolor{cmtgreen}{The ground truth shape of the rope was obtained by locating the left control point (end-effector), right fixed endpoint and the middle three Aruco Markers shown \cite{Ramirez_2018_Aruco} in the Fig. \ref{fig:Cover_Photo} and Fig. \ref{fig:real2sim}. We fixed the simulated rope to the same right endpoint. By setting the middle three points as the shape target, we optimized the position of the control point from the differentiable \textcolor{cmtorange}{Thomas} XPBD simulation.}

As shown in Fig. \ref{fig:real2sim}, the transparent rope \textcolor{cmtgreen}{with low opacity} was our \textcolor{cmtred}{shape control} target status, and the solid ones \textcolor{cmtgreen}{with high opacity} were the result \textcolor{cmtred}{controlled by the identified parameters from Section \ref{sec:inextensible_para_identi_result}. Our controlled shape could almost overlap with the target rope, as shown in the \textbf{Top Row} of Fig. \ref{fig:real2sim}. However, some had deviations as shown in the \textbf{Bottom Row} of Fig. \ref{fig:real2sim}.} \textcolor{cmtgreen}{One of the reason is we only considered three shape points as the target for loss computation. The accuracy will be improved by considering more discretized segments of the ground truth data. Another reason} was because we did not consider the \textcolor{cmtgreen}{rotational control} of the end-effector when solving the inverse kinematics of the Baxter, which caused the rope not to be able to move to the desired position \textcolor{cmtgreen}{perfectly}.

\section{Conclusion and Future Works}
\textcolor{cmtred}{
This paper used a compliant position-based framework to conduct the differential real-to-sim tasks for parameter identification and shape control tasks. Several geometrical constraints were introduced to model the rope-like objects' coupling stretch/shear and bending/twisting effects. To inspect the inextensible and extensible impact, the Thomas solver and the Jacobi solver are proposed for the distance constraint. The experiment results on the Baxter robot and DVRK platform proved the validity and robustness of our solvers. The shape control tasks showed a novel path for real-to-sim robotic manipulation operations.
}

\textcolor{cmtred}{
The future works will consider differentiable control with collision handling and rigid-deformable coupling. Meanwhile, more advanced tasks will be considered, such as surgical thread manipulation in field environments using the proposed inextensible solver.
}

\section*{Acknowledgement}
Many thanks to Yutong Zhang for the rendering of the cover photo. This work was supported by NSF CAREER award $\#$2045803 and the US Army Telemedicine and Advanced Technologies Research Center.


{\small
\bibliographystyle{ieeetran}
\bibliography{IEEEcitation}
}

\end{document}